\documentclass[journal]{IEEEtran}
\usepackage{amsmath}
\usepackage[ruled,vlined,linesnumbered]{algorithm2e}
\usepackage{xcolor}
\usepackage{multirow}
\usepackage{booktabs}
\usepackage{tabularx}
\usepackage{amsmath}
\usepackage{amsfonts}
\usepackage{graphicx}
\usepackage{adjustbox}
\usepackage{threeparttable}

\usepackage{pdfpages}

\definecolor{light-gray}{gray}{0.85}

\newcommand{\highlight}[1]{\colorbox{light-gray}{\color{black}\textit{#1}}}

\newcommand{\sm}[1]{\textcolor{black}{#1}}

\newcommand{\rev}[1]{\textcolor{black}{#1}}
\newcommand{\revt}[1]{\textcolor{black}{#1}}

\begin{document}
\title{A Novel Generalised Meta-Heuristic Framework for Dynamic Capacitated Arc Routing Problems}

\author{Hao Tong, \IEEEmembership{Member,~IEEE}, Leandro L. Minku, \IEEEmembership{Senior Member, IEEE}, Stefan Menzel, Bernhard Sendhoff, \IEEEmembership{Fellow, IEEE}, and Xin Yao, \IEEEmembership{Fellow, IEEE}
   \thanks{Hao Tong, Leandro L. Minku and Xin Yao (\emph{the corresponding author}) are with the School of Computer Science, University of Birmingham, Edgbaston, Birmingham B15 2TT, UK. (email: hxt922@cs.bham.ac.uk, L.L.Minku@cs.bham.ac.uk, xiny@sustech.edu.cn)} 
   \thanks{Stefan Menzel \sm{and} Bernhard Sendhoff are with the Honda Research Institute Europe GmbH, 63073 Offenbach, Germany. (email: stefan.menzel@honda-ri.de, bernhard.sendhoff@honda-ri.de)}
   \thanks{Xin Yao (\emph{the corresponding author}) is also with the Research Institute of Trustworthy Autonomous Systems (RITAS), Department of Computer Science and Engineering, Southern University of Science and Technology, Shenzhen 518055, China.}
}

\maketitle


\begin{abstract}
The capacitated arc routing problem (CARP) is a challenging combinatorial optimisation problem abstracted from \rev{many} real-world applications, \rev{such as} waste collection, \rev{road gritting} and mail delivery. However, few studies considered dynamic changes during the vehicles' service, which can \rev{cause} the original schedule infeasible or obsolete. The few existing studies are limited by the dynamic scenarios \rev{considered}, and by \rev{overly complicated} algorithms that \rev{are} unable to benefit from the wealth of contributions provided by the \rev{existing} CARP literature. \rev{In this paper}, we \rev{first} provide a mathematical formulation \rev{of} dynamic CARP (DCARP) \rev{and design a simulation system that is able to consider dynamic events while a routing solution is already partially executed}. We then propose a novel framework \rev{which can benefit from} existing static CARP optimisation algorithms so that they 
could be used to handle DCARP instances. The framework \rev{is very flexible. In response to a dynamic event, it can use either a simple restart strategy or} a sequence transfer strategy that benefits from past optimisation experience. Empirical studies \rev{have been} conducted on a wide range of DCARP instances \rev{to evaluate our proposed framework}. The results show that the proposed framework significantly improves over \rev{state-of-the-art dynamic optimisation} algorithms.
\end{abstract}


\section{Introduction}
The Capacitated Arc Routing Problem (CARP) is a classical combinatorial optimisation problem with a range of \revt{collection and delivery} applications in the real world.
\revt{For example, in a waste collection problem \cite{lacomme2005evolutionary}, the capacitated vehicles start from a depot to collect the waste distributed in different streets. In a winter road gritting problem, which is a kind of delivery application \cite{handa2006robust}, the fully loaded vehicles deliver the salt to spread into different required roads. Such scenarios are the main focus in this paper.}

Constructive heuristic methods, such as Ulusoy's split \cite{ulusoy1985fleet} and Path-Scanning \cite{golden1983computational}, were proposed to construct feasible executable solutions for CARP based on an optimised sequence of tasks. Tabu search \cite{brandao2008deterministic}, memetic algorithms \cite{tang2009memetic} and others were also proposed to \rev{solve the} CARP.
In addition, efficient algorithms have been proposed to tackle large scale \rev{CARPs} \cite{tang2016scalable, whlk2018fast}. \rev{Many} different variants of CARP have been investigated \cite{mouro2017update,corbern2020arc}. For example, multi-depot CARP considers several different depots in the graph \cite{xing2009evolutionary}, and open CARP allows the routes to be open with different starting and ending nodes \cite{arakaki2018hybrid}. Split-delivery CARP \rev{allows} the edge demand \rev{to} be served by several vehicles \cite{belenguer2010split}. Periodic CARP \rev{considers the cases} where the tasks are required to be served with a certain number of times over a given multiperiod horizon \cite{chen2018two}. Time CARP \rev{considers} the time instead of the volume restriction of the vehicles \cite{dearmas2018solving}.

However, all these studies concentrate on static \rev{CARPs}, where the problem \rev{remains static during the entire time of a solution's execution}. In real applications, dynamic changes usually happen when vehicles are in service, \rev{i.e., when a solution is partially executed,} thus influencing the vehicles' \rev{follow-on} service. For example, a road may be closed due to an accident or new tasks may emerge during the vehicles' service. When that happens, a new graph, i.e. a new problem instance, is formed, in which vehicles would stop at different locations, labelled as outside vehicles, with various amounts of remaining capacities. \rev{As a result, the current schedule may become inferior or even feasible}. \rev{Dynamic CARP (DCARP) in our paper thus aims at re-scheduling the service plan \cite{liu2014memetic, liu2014benchmark}}.
For clarity, the following three different concepts are used throughout our paper:
\begin{itemize}
    \item \textbf{DCARP:} A variant of CARP where the status of a graph is changed due to \sm{dynamic events occurring} during a CARP solution's execution.
    \item \textbf{DCARP Instance:} The updated graph with some outside vehicles after the dynamic events happen.
    \item \textbf{DCARP Scenario:} \rev{A} scenario contains a series of DCARP instances with the whole service process, starting from executing an initial solution in the original CARP map until all tasks are served.
\end{itemize}
\rev{It is worth noting that Mei et al. \cite{mei2013evolutionary} used the term DCARP to denote the uncertain CARP, which is different from the meaning of DCARP in this paper. 
Uncertain CARP focuses on robust optimisation \cite{mei2013evolutionary,liu2021robust}, where one is interested in finding solutions that are \textit{robust} to uncertainties, such as changing degrees of congestion or level of demands. However, there are dynamic events in the real world which can not be handled well by robust optimisation, such as closure of roads or addition of new tasks. As a result, dynamic optimisation, which is the focus of our paper, has been an active research topic in recent years.}

DCARP, \rev{as we defined early}, was \rev{first} investigated in \cite{handa2005dynamic} when considering the salting route optimisation problem\rev{. However} few studies in the literature have focused on DCARP so far. Mariam et al. \cite{tagmouti2011dynamic} solved DCARP with time-dependent service costs motivated from winter gritting applications. Liu et al. \cite{liu2014memetic} defined some \rev{ dynamics} in DCARP and proposed a benchmark generator for DCARP \cite{liu2014benchmark}. Marcela et al. \cite{monroy2017rescheduling} \rev{dealt} with rescheduling for DCARP, which considered the failure of vehicles. Wasin et al. \cite{padungwech2020effects} considered new tasks in DCARP. A robot path planning problem \cite{yazici2009dynamic} and our previous work \cite{tong2020towards} focused on the split scheme in DCARP. \sm{Split} schemes convert an ordered task sequence into an executable solution \rev{of} multiple explicit routes.

Even though DCARP has been \rev{investigated by different people}, there \rev{is still a lack of formal} mathematical formulation of DCARP to the best of our knowledge. 
\rev{The field also lacks a system that can simulate the behaviour of vehicle's service process in the real world.}
\rev{Liu et al. \cite{liu2014benchmark}} proposed a benchmark generator for DCARP. However, their generator \rev{cannot consider dynamic events during the execution of a routing solution} and, thus, is unsuitable for our DCARP scenarios, \rev{where changes happen during the execution of the scenarios.}
\rev{Finally, there is a rich literature on existing CARP optimisation algorithms that could potentially contribute towards DCARP optimisation, but they are not applicable to DCARP instances. This is because they work under the assumption that all vehicles start at the depot and have the same capacities, which is not the case in DCARP. A framework to enable the application of existing CARP optimisation algorithms to DCARP problems is desirable.}

\rev{Therefore, this paper has the following contributions:}
\begin{enumerate}
    \item \rev{We provide the first mathematical formulation of DCARP in the literature.}
    \item \rev{We design a simulator to simulate the behaviour of vehicles' service processes in the real world. The simulator is developed according to the collection or gritting problem, where the vehicle does not have to return to the depot for \revt{loading new different delivered items}. \revt{It offers a} novel research platform to support DCARP studies.}
    \item \rev{We propose a novel framework capable of generalising almost all existing algorithms designed for static CARP to the DCARP context.
    The framework converts a DCARP instance into a ``static'' CARP instance by introducing the idea of ``virtual tasks'', which enables outside vehicles (with potentially partial capacity) to be interpreted as vehicles located at the depot (with their full capacity). The DCARP instance can then be solved as if it was a static CARP instance by static CARP algorithms. After a solution is found, its corresponding DCARP route where the vehicles start at their outside positions is generated.}
    \item \rev{As a dynamic scenario is composed of a series of DCARP instances, \rev{similarities} between DCARP instances \rev{can and should be exploited}. Therefore, we propose two strategies for generating initial solutions in our framework, namely a \emph{sequence transfer strategy} and a \emph{restart strategy}, to solve \rev{a} new DCARP instance.} The sequence transfer strategy generates a potentially good solution based on the previous optimisation experience by transferring the sequence of remaining unserved tasks. The restart strategy starts from scratch without using any information and optimises each DCARP instance independently of each other.
    \item \rev{We perform extensive experiments with a variety of DCARP instances, demonstrating the effectiveness of the proposed framework. We show that valuable research progress achieved by the static CARP literature can contribute towards optimisation results that significantly outperform the existing algorithm \cite{liu2014memetic} that was specifically designed for DCARP.}
\end{enumerate}

The remainder of this paper is organised as follows. Section \ref{related-work} discussed the related work on DCARP and this paper's motivation. After that, a general mathematical formulation of DCARP and a simulation system for DCARP are provided in Section \ref{formulation}. Section \ref{framework} introduces the main algorithm of our generalised optimisation framework for DCARP. Section \ref{experiment} presents \rev{our experimental} study on the proposed framework to evaluate its \sm{efficiency}. Section \ref{conclusion} concludes the paper.

\section{Related work and motivation}\label{related-work}
\rev{In the literature, there are two related but different research topics,} which target the (re)scheduling of vehicles in dynamic environments: Dynamic CARP (DCARP) and dynamic vehicle routing problem (DVRP). DCARP focuses on serving \rev{tasks which are the arcs in the graph} while DVRP focuses on serving vertices. With respect to DCARP, few approaches \rev{were} proposed.
Liu et al. \cite{liu2014memetic} proposed a memetic algorithm with a new distance-based split scheme (MASDC) for DCARP. However, its performance is \rev{unsatisfactory} since it suffers from noise in the fitness evaluation due to the impact of random splits, as well as the neglecting available vehicles placed in the depot. Monroy et al. \cite{monroy2017rescheduling} considered \rev{only} the broken down vehicles and presented a heuristic to minimise the operations and disruption cost. Padungwech et al. \cite{padungwech2020effects} considered \rev{only} the new tasks during the vehicles' service. They applied tabu search to optimise the DCARP, in which the solution is represented as routes with different start vertices.

As stated above, DVRP focuses on serving vertices, instead of tasks/arcs (i.e., arcs with demands), and its research work mainly comprises two categories called dynamic deterministic VRP and stochastic VRP according to if problem knowledge is used during the optimisation or not \cite{pillac2013review}.
For solving DVRP, it might be possible to transform DVRP instances into DCARP instances or vice versa, \rev{such transformation will increase} the problem's dimension \cite{foulds2014compact}. \rev{For example, the number of vertices in the CVRP instance will increase if transformed from a CARP instance. The number of vertices to be served will be greater than the number of arcs to be served. Furthermore, dynamics events in VRP and CARP are very different. In short, it is not a suitable approach to convert CARP into capacitated VRP (CVRP) and then solve CVRP. It is better to design CARP or DCARP specific algorithms as the research community has been doing for many years.}

There are two representations commonly used in optimisation \rev{algorithms} for CARP in the literature. The first type provides all explicit routes in the solution, separated by a dummy task \cite{tang2009memetic}, while \rev{the other} type is an ordered list of tasks without separation. These two types of \rev{representation can be} used together in the algorithm for CARP. For example, constructive heuristics, such as Path-Scanning \cite{golden1983computational} generates solutions with explicit routes. This representation is \rev{friendly to} local search operator. The representation with an ordered list of tasks is \rev{often} used in meta-heuristic algorithms with crossover operators, \rev{such as} memetic algorithms \cite{ lacomme2005evolutionary,tang2009memetic}. The Ulusoy's split scheme \cite{ulusoy1985fleet} is an exact algorithm for converting an ordered list of tasks to a solution with explicit routes by building an auxiliary graph according to the task sequences. 

\rev{For DCARP,} the calculation \rev{of} cost and capacity violation of the routes corresponding to the outside vehicles are required to be specifically considered due to the fact that outside vehicles have different locations and remaining capacities. 
\rev{It is more complicated} to use an ordered list of tasks as the solution representation during the optimisation because the Ulusoy's split \cite{ulusoy1985fleet} is not suitable anymore and specific split schemes \rev{\cite{tong2020towards}} are required. Even though \rev{a new split scheme was} proposed in our previous work \cite{tong2020towards}, \rev{its high computational complexity limits its performance}. \rev{Therefore, the existing algorithm for static CARP \cite{liu2014memetic} is adapted to solve DCARP instance with some modification, such as the existing work in \cite{liu2014memetic}. }

In this paper, we \sm{propose a novel \rev{general} framework, which enables the adoption} of existing CARP algorithms to DCARP. \rev{However, this does not exclude future development of new dedicated dynamic algorithms.} In the next section (Section \ref{formulation}), we \rev{will} introduce \rev{our} mathematical formulation \rev{of DCARP} and a newly designed simulation system, followed by our \rev{general} framework in Section \ref{framework}.

\begin{table}[htbp]
  \centering
  \caption{Glossary of Mathematical Notations Used in this paper} \label{notations}
    \begin{tabular}{cl}
    \toprule
    Symbols & Meaning \\
    \midrule
    $G$ & Graph $G=(V, A)$\\
    $V$  & Set of vertices. \\
    $A$  & Set of arcs. \\
    \rev{$I_m$} & \rev{The $m^{th}$ DCARP instance.} \\
    $v_0$ & The depot. \\
    $dm(u)$ & The demand of an \rev{arc $u\in A$ (ID)}. \\
    $dc(u)$ & The deadheading(traversing) cost of an arc \rev{$u\in A$ (ID)}. \\
    $sc(u)$ & The serving cost of an arc \rev{$u\in A$ (ID)}. \\
    \rev{$state(u)$} & \rev{The traffic property of an arc $u \in A$ (ID)}. \\
    $N_t$ & The number of \rev{tasks, $N_t =|R|$.} \\
    \rev{$N_{veh}$} & The maximum number of vehicles. \\
    $Q$  & The capacity of empty \sm{vehicles}. \\
    $OV$ & The set of outside vehicles. \\
    $N_{ov}$ & The number of outside vehicles, \rev{$N_{ov} = |OV|$}. \\
    $q_k$ & The remaining capacity of the \rev{$k^{th}$} outside vehicle. \\
    $mdc(v_i, v_j)$ & The minimal total deadheading cost from vertex $v_i$ to $v_j$. \\
    $head_{t}$ & The head node of task $t$. \\
    $tail_{t}$ & The tail node of task $t$. \\
    $S$  & A DCARP solution, \rev{i.e., a set of routes}. \\
    $r_k$ & The $k^{th}$ route.  \\
    $l_k$ & The number of tasks in $k^{th}$ route. \\
    \rev{$t_{k, i}$} & \rev{The $i^{th}$ task in $k^{th}$ route.} \\
    \rev{$RC_{r_k}$} & \rev{The total cost of $k^{th}$ route $r_k$.} \\
    $TC(S)$ & \rev{The total cost of solution $S$}. \\
    \bottomrule
    \end{tabular}%
    \begin{tablenotes}
        \item \rev{The set of outside vehicles $OV$ depends on the dynamic changes that different changes at different times will lead to different $OV$.}
    \end{tablenotes}
\end{table}%

\section{Problem Formulation and Simulation System}\label{formulation}
In this section, we provide the first mathematical formulation for DCARP. The mathematical notations used in this paper are summarised in Table \ref{notations}. A new simulation system is then \rev{proposed} to generate benchmark \rev{instances from the} existing CARP benchmark for testing DCARP algorithms.

\subsection{Notations and Mathematical Formulation}
For simplicity, in the present paper we consider the collection or gritting application, i.e., vehicles can continue the service paths without requiring to return to the depot \revt{when new tasks/demands appear.} \rev{Such} DCARP scenario is composed of a series of DCARP instances: $\mathcal{I} = \{I_0,I_1, ..., I_m, ..., I_M\}$. Each DCARP instance corresponds to a problem state, which contains all the information regarding the state of the map and vehicles involved in the routing problem, and highly depends on the previous instance and the solution's execution. The initial problem instance $I_0$ is a conventional static CARP, in which all vehicles are located at the depot \rev{having the same full} capacities. We can obtain an initial solution \rev{in $I_0$} and execute this solution in the graph. During the execution, some \rev{dynamics} \cite{liu2014memetic} happen at random \rev{points in time} when vehicles are in service, thus changing the problem instance and \rev{potentially} requiring a new \rev{better} solution. Vehicles then continue to serve tasks from the positions they had stopped (stop points). DCARP terminates when all tasks are served, and all vehicles have returned to the depot. In a DCARP scenario, the key objective is to achieve a schedule cost, which \rev{should be as} low as possible for each DCARP instance. Let's first focus on the mathematical formulation \rev{for} one DCARP instance.

The map for any DCARP instance $I_m$ is \rev{provided} as a graph $G$. Suppose the map of a DCARP instance $I_m$ is represented by $G=(V, A)$ with a set of vertices $V$ \rev{and} arcs (directed links) $A$. There is a depot $v_0 \in V$ in the graph, which contains vehicles that are not yet serving any tasks. The \sm{set} $A$ is \rev{given} by
\begin{equation*}
    A = \{ <v_i, v_j> | v_i, v_j \in V \} \\
\end{equation*}
where for each arc \rev{$u$, i.e. $<v_i, v_j> \in A$}, $v_i$ is the head vertex and $v_j$ is the tail vertex. A given arc $<v_i, v_j>$ only exists if it is possible to traverse from vertex $v_i$ to vertex $v_j$ without passing through other vertices. 
Each arc $u$ in the graph is associated with a deadheading \rev{(traversing)} cost \rev{$dc(u)$}, a serving cost \rev{$sc(u)$} and a demand \rev{$dm(u)$}. The deadheading cost of an arc means \rev{the cost} that the vehicle just traverse \rev{this arc} without serving while the serving cost is the cost when vehicles serve this arc. \rev{For simplicity, the deadheading cost are assumed to be symmetric in this paper.} The deadheading cost has been included in the serving cost such that the deadheading cost is not required to be calculated when the vehicle serves an arc.
A subset \rev{$R \subseteq A$} contains all arcs required to be served in the graph. The arc $u \in R$ is named as `task` and \rev{has} a positive demand $dm(u) > 0$. \rev{For convenience, we use $t$ to represent a task, and use an arc ID for identification.} 

\sm{The} DCARP instance $I_0$ only contains vehicles at the depot. As for DCARP instances \rev{$I_m$($m>0$)}, in addition to vehicles that are currently at the depot, there may also be outside vehicles with remaining capacities. These are vehicles that had already started to serve tasks when \rev{a dynamic event occurs}. 
Suppose there are $N_{veh}$ vehicles in total with a maximum capacity $Q$ at the depot and $N_{ov}$ ($N_{ov} \leq N_{{veh}}$) outside vehicles with remaining capacities $\{ q_1, q_2, ..., q_{N_{ov}}\}$. The stop points (locations) \sm{of the} outside vehicles are \sm{labelled} as $OV = \{v_1, v_2, ..., v_{N_{ov}}\}$.
The optimisation of DCARP aims to reschedule the \rev{remaining tasks with the minimal} cost considering both outside and depot vehicles.. 

A DCARP solution $S = \{ r_1, r_2, ..., r_{N_{ov}}, ..., r_K \}$ contains $K$ routes\sm{, where the routes} $r_1$ to $r_{N_{ov}}$ start from \rev{locations that outside vehicles located} while routes $r_{N_{ov}+1}$ to $r_{K}$ start from the depot. \rev{Each route can be represented by three components: starting vertex, an ordered list of tasks (arc IDs) and the final depot.} \rev{Therefore, a} given route $r_k$ can be expressed as $r_k = \rev{(}v_k, t_{k, 1}, t_{k, 2}, ..., t_{k, l_k}, v_0\rev{)}$, where the vehicle starts from stop location $v_k$ and returns to the depot $v_0$, \rev{whereas} $l_k$ denotes the number of tasks served by route $r_k$. For route $r_k$, where $k > N_{ov}$, $v_k$ equals to $v_0$. \rev{This representation is very easy to be converted to an explicit route by connecting two subsequent tasks using Dijkstra’s algorithm so that the route cost can be calculated.} In addition, a DCARP solution has to satisfy three constraints which are the same as constraints in static CARP:
\begin{itemize}
    \item Each route served by one vehicle must return to \sm{the} depot.
    \item Each task has to be served once.
    \item The total demand for each route served by one vehicle cannot exceed the vehicle's capacity $Q$.
\end{itemize}
Due to the different remaining capacities for outside vehicles, the capacity constraint is required to be formulated for each outside vehicle separately. As a result, the objective function \rev{and the constraints} for DCARP \rev{are} given as follows:
\begin{equation} 
    \label{objective1}
    \begin{aligned}
        \mathbf{Min} &\quad TC(S) = \sum_{k=1}^{K} RC_{r_k}  \\
        \mathbf{s.t.} &\quad \sum_{k=1}^{K}l_k = N_t \\
        &\quad t_{k_1, i_1} \neq t_{k_2, i_2}, \rev{\mathrm{for\ all\ }} (k_1, i_1) \neq (k_2, i_2) \\
        &\quad \sum_{i=1}^{l_k}dm(t_{k, i}) \leq q_k, \forall k \in \{1, 2, ... N_{ov}\}  \\
        &\quad \sum_{i=1}^{l_k}dm(t_{k, i}) \leq Q, \forall k \in \{ N_{ov}+1, ..., K\} 
    \end{aligned}
\end{equation}
where $N_t$ is the number of tasks and $RC_{r_k}$ \sm{denote} the total cost of route $r_{k}$ and is computed according to Eq. \ref{objective2}:
\begin{equation}
    \label{objective2}
    \begin{aligned}
        RC_{r_k} &= mdc(v_k, tail_{t_{k, 1}}) + 
        mdc(head_{t_{k, l_k}}, v_0) + \\
        &\sum_{i=1}^{l_k-1}mdc(head_{t_{k, i}}, tail_{t_{k, i+1}}) + \rev{\sum_{i=1}^{l_k}sc(t_{k, i})}
    \end{aligned}
\end{equation}
where $head_{t}$, $tail_{t}$ denotes the head and tail vertices of the task, $mdc(v_i, v_j)$ denotes the minimal total deadheading cost traversing from node $v_i$ to node $v_j$, \rev{and $sc(t_{k, i})$ denotes the serving cost of task $t_{k, i}$ }. The first two constraints in Eq. (\ref{objective1}) guarantee that all tasks are served only once and the other two constraints are formulated to satisfy the capacity constraint. 

\subsection{Simulation System for DCARP}\label{simulation-system}

\begin{figure}[!htpb]
    \centering
    \includegraphics[width=0.9\linewidth]{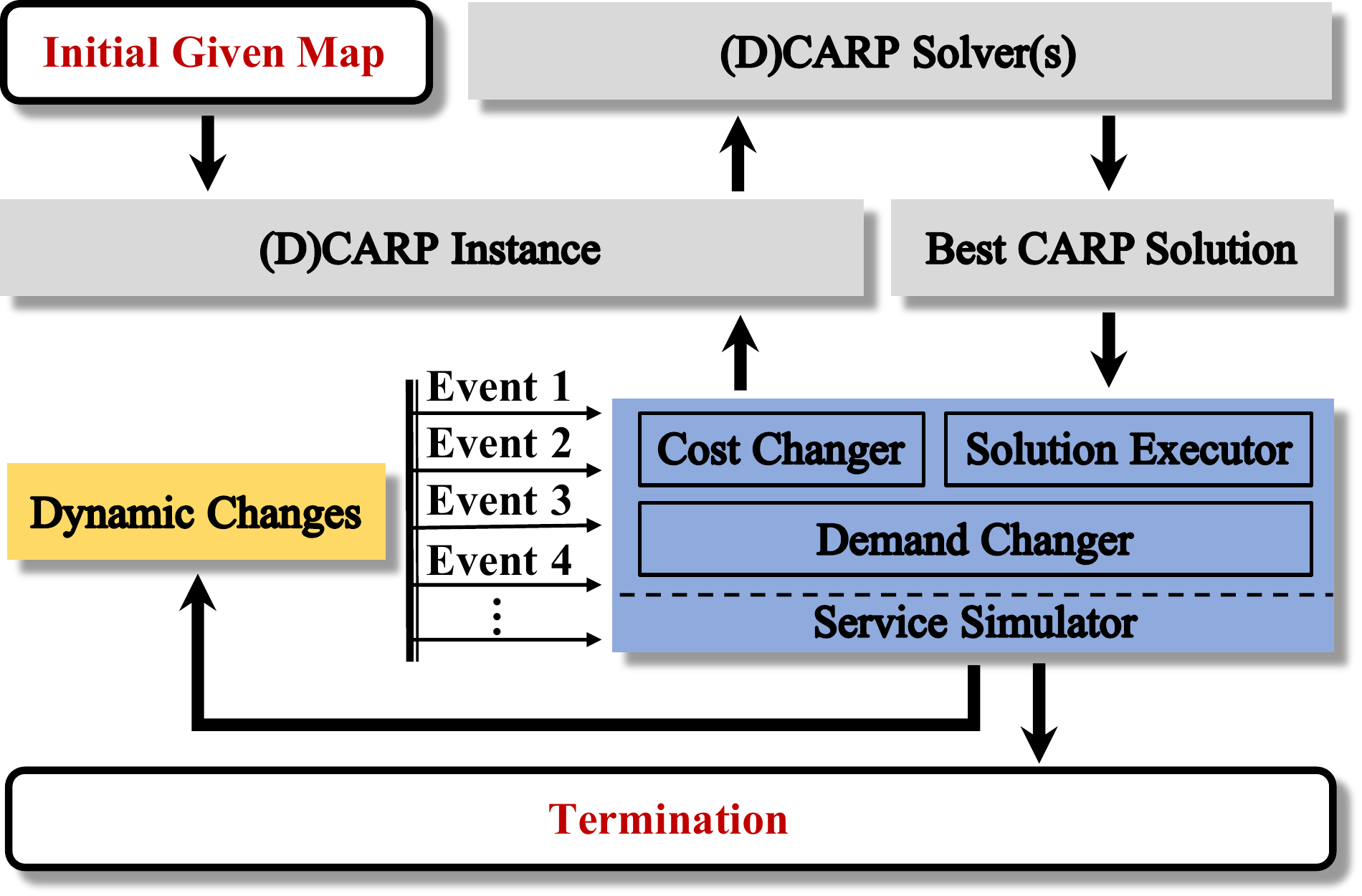}
    \caption{The architecture of \sm{our} simulation system.} \label{simulator}
\end{figure}

In order to test optimisation algorithms for DCARP, a simulation system that includes some \sm{common} dynamic events is required.
Even though a benchmark generator for DCARP has been proposed by Liu et al. \cite{liu2014benchmark}, it \rev{has} shortcomings, which prevent it to be used as research platform.
Intuitively, a DCARP instance should be generated \rev{from the dynamic change of a previous DCARP instance during a solution's execution}, such as road congestion or recovering from the congestion. However, these essential details are not considered in the existing benchmark generator \cite{liu2014benchmark}. Therefore, we \rev{have} designed a simulation system which includes nine commonly occurring events and generates DCARP instances from the existing CARP benchmark\footnote{\rev{https://github.com/HawkTom/Dynamic-CARP}}. Nine events and their corresponding \rev{changes in} mathematical forms are listed in Table \ref{dynamic-change}, and the simulation system\rev{'s architecture} is presented in \rev{Figure \ref{simulator}}.

\begin{table}[!htpb]
    \caption{Types of dynamic events in DCARP. The events with * are new events considered in this paper, \rev{which have never been considered in the literature.}}\label{dynamic-change}
    \centering
    \renewcommand{\arraystretch}{1.5}
    \setlength{\tabcolsep}{5pt}
    \color{black}{
    \begin{tabular}{ll}
    \toprule
    Event types                     & Changes \\ \midrule
    \multirow{2}{*}{1. Vehicle break down}   & Collection: $ dm(u_k): 0 \rightarrow Q-q_k$                       \\
       & Delivery: $dm(u_k):$ No change\\
    2. Road closure      & $ dc(u): \overline{dc(u)} \rightarrow \infty  $                \\
    3. Congestion        & $dc(u): \overline{dc(u)} \rightarrow \overline{dc(u)} + c $            \\
    4. Recover from roads closure * & $dc(u): \infty \rightarrow \overline{dc(u)} $                \\
    5. Recover from congestion *  & $dc(u): dc'(u) \rightarrow \overline{dc(u)} $   \\
    6. Congestion become worse * &  $dc(u): dc'(u) \rightarrow dc'(u)+c $  \\
    7. Congestion become better * &  $dc(u): dc'(u) \rightarrow dc'(u)-c $  \\
    8. Demand increases      & $dm(u): dm(u) \rightarrow dm(u) + d $ \\
    9. Added tasks           & $dm(u): 0 \rightarrow d $      \\ \bottomrule
    \end{tabular}
    \begin{tablenotes} 
	    \item $\overline{dc(u)}$ is the expected deadheading cost of arc $u$ without congestion. ${dc'(u)}$ is the deadheading cost of arc $u$ with congestion.  $c, d$ are the changing cost and demand, respectively. Event 5 is a special case of Event 7 where $c = dc'(u) - \overline{dc(u)}$ in Event 7.
     \end{tablenotes}
    }
    \vspace{0.5em}
\end{table}

\rev{To make our simulator more close to realistic events, we} have added several dynamic events, which have not been considered in the literature. \rev{For example}, the road can recover from a closure or a congestion, which has been marked with a star (*) in Table \ref{dynamic-change}.
If a vehicle breaks down, we can assume that this vehicle $k$ has already served some tasks. Consequently, the demand of the arc $u_k$, where \rev{the vehicle $k$ broke down}, increases from $0$ to $Q-q_k$ to include the \rev{already served} loads \rev{in collection applications}. For \revt{delivery} applications, the broken down vehicle has no impact to the demand of the arc.
\revt{As we mainly considered collection or gritting problems, we assumed in our work that the tasks/demands will not vanish until fully served, although this may be extended and addressed in future. However, our proposed simulator and framework is still applicable and capable of taking these events into account in case they would be added in future research.} Based on Table \ref{dynamic-change}, we can easily observe that all dynamic events impact the cost or demand of arcs. Therefore, we designed the \textit{cost changer} and \textit{demand changer} in our simulation system to simulate these events (Figure \ref{simulator}).

\begin{algorithm}[!h]
    \caption{The pseudo code of the service simulator}\label{service-simulator}
    \KwIn{Executable solution $S$, Previous instance $I_m$}
    \KwOut{\rev{The new instance $I_{m+1}$}}
    \vspace{0.5em}
     Set probabilities of occurrence for all events: $[p_{event}, p_{road}, p_{bdrr}, p_{crr}, p_{crbb}, p_{icd}, p_{add}]$; \label{l1} \\
     
     \vspace{0.3em}
     Execute $S$ on $I_m$; \label{l2-1} \\
     Select a uniformly random time to stop execution; \\ 
     Remove all served tasks: make demand of all served tasks be $0$; \label{l2-2} \\
     \textbf{Event 1} Randomly select $n$ vehicles to break down. \label{l3-1} \\ 
     \textbf{/**** Cost Changer ****/} \\
     
     \For{each arc $u_i \in A$}
     {  \label{l4-1}

        \If{$rand() < p_{event}$}{
            \Switch{$state(u_i)$}{
            \Case{0}{
                \If{$rand() < p_{road}$}{
                \textbf{Event 2}  $dc(u_i) = \infty$, 
                $state(u_i) = 2$\\
                }\Else {
                \textbf{Event 3} $dc(u_i) = \overline{dc(u_i)}+c$,  
                $state(u_i) = 3$
                }
                \break
            }

            \Case{2}{
                \If {$rand() > p_{bdrr}$}{
                    \textbf{Event 4} $dc(u_i) = \overline{dc(u_i)}$, 
                    $state(u_i) = 0$
                }
                \break
            }

            \Case{3}{
                $r = rand()$\\
                \uIf{$r < p_{crr}$}{
                    \textbf{Event 5} $dc(u_i) = \overline{dc(u_i)}$, 
                    $state(u_i) = 0$
                }
                \uElseIf{$r < p_{crbb}$}{
                    \textbf{Event 6} $dc(u_i) = dc'(u_i) + c$
                }
                \Else{
                    \textbf{Event 7} $dc(u_i) = dc'(u_i) - c$
                }
                \break
            }
        }
        }
    }\label{l4-2}
      \textbf{/**** Demand Changer ****/ }\\
    \For{each arc $u_i \in A$}
    { \label{l5-1}
        \If{$dm(u_i) > 0$ and $rand() < p_{icd}$}{
            \textbf{Event 8} $dm(u_i) = dm(u_i) + d $
        }
        \If{$dm(u_i) == 0$ and $rand() < p_{add}$}{
            \textbf{Event 9} $dm(u_i) = 0 + d $
        }
     }\label{l5-2}

\end{algorithm}

In Figure \ref{simulator}, the system starts from a provided initial graph, which could be taken from existing benchmarks for static CARP. Then, \revt{a CARP solver selected by the user}, such as \rev{solvers based on} memetic algorithms \cite{tang2009memetic}, can be used to obtain the initial solution, i.e. the first schedule for the vehicles. The core part of the system is the \emph{Service Simulator} in Figure \ref{simulator}, which is used to execute the CARP solution and update the graph. The pseudocode \rev{for the \emph{Service Simulator}} is presented in Algorithm \ref{service-simulator}. During the execution of the CARP solution, the maximum number of vehicles in the depot \rev{($N_{veh}$)} is considered. If the number of routes in the schedule \sm{exceeds} the predefined maximum number of vehicles, the route with the smallest cost will be served first and the remaining routes will be served after some vehicles return to the depot.
When the solution is executed, some dynamic \rev{events} will happen \rev{according to a series of predefined parameters listed in Line \ref{l1}} and \revt{influence the graph at a uniformly random time between the service start and completion time.} \rev{After that, a new DCARP instance is generated based on the dispatched solution. Different solutions will result in different DCARP instances, which \rev{might not facilitate} fair comparison of different algorithms. Therefore, we apply the best solution among all solutions obtained by different algorithms to the service simulator and generate one new DCARP instance for all algorithms for the fair comparison.}

Once the dynamic change happens \rev{on the DCARP instance $I_m$}, the \emph{Service Simulator} stops the execution of the current solution. \rev{Then,} the \textit{cost changer} and \textit{demand changer} will update the DCARP instance. First, as broken down vehicles influence only a specific arc, we simulate Event 1 separately from other dynamic events. The algorithm randomly selects $n$ vehicles \rev{from all dispatched vehicles} to break down, as shown in Line \ref{l3-1}. Events 2 to 7 will influence the cost of several arcs so that the \textit{cost changer} mainly simulates these five events, as shown in Lines \ref{l4-1}-\ref{l4-2}. Each arc $u_i$ has a \rev{traffic} property, $state(u_i)$, recording whether it is currently in a changed state \rev{compared to its original state having expected cost without traffic events}.

\rev{In the \textit{cost changer}, the simulator firstly determines whether or not a change from the current state occurs according to the probability $p_{event}$ for each arc. If a change occurs, a dynamic event is triggered according to the events' probabilities and arc's current changed state. If an arc keeps the original state, i.e. $state(u_i) == 0$, Event 2 or 3 happens in this arc depending on probability $p_{road}$.}
If $state(u_i) == 2$, the road has broken down before so that it recovers with \sm{a} probability $p_{bdrr}$. If $state(u_i) == 3$, the road is in congestion. It may either \rev{completely} recover with \sm{a} probability $p_{crr}$, or the traffic jam may ease or get worse (by a random cost) with \sm{a} probability $p_{crbb}$ or $1 - p_{crbb}$, respectively. Compared to Events 2 to 7, Events 8 and 9 \rev{in the \textit{demand changer}} are much easier to implement, because we assume that the tasks/demands do not vanish unless served completely and the demand of tasks can only increase \revt{under the practical scenarios considered in this paper}. Event 8 may happen to a task with a probability $p_{icd}$, increasing the demand by a random amount. For arcs with no demand, Event 9 will happen with \sm{a} probability $p_{add}$.

Finally, we will get a new DCARP instance \rev{$I_{m+1}$}, and the solver generates a new DCARP solution. The system terminates after all tasks are served.

\section{A Generalised Optimisation Framework for DCARP}\label{framework}
In this section, we propose a virtual-task strategy to change a DCARP instance to a `virtual static' instance. After that, a generalised optimisation framework based on a virtual-task strategy for DCARP with two different initialisation strategies is proposed, which \rev{can make use of} algorithms for static CARP for solving DCARP.

\subsection{Virtual task}
As discussed in the previous section, the main challenge of scheduling vehicles for DCARP by using algorithms designed for static CARP is to take the outside vehicles with different locations and remaining capacities into account. 
We propose a virtual task strategy that forces all outside vehicles to virtually return to the depot \rev{for optimisation purposes,} such that all vehicles \rev{(some virtually) start} at the depot \rev{during the optimisation. As a result,} algorithms for static CARP, which assume that all vehicles start at the depot, can be adopted. \rev{After the optimisation, the obtained solution with routes starting from the depot will be converted to an executable solution according to the locations of outside vehicles. In other words, even though the outside vehicles will virtually return to the depot for running the optimisation process, in the executable solutions themselves, the outside vehicles start their new routes from their outside locations. For this strategy to work, some adjustments need to be made so that the optimisation problem with virtual tasks is equivalent to the actual DCARP instance being solved. Such adjustments will be explained next.}

\rev{The pseudocode of constructing the virtual task is presented in Algorithm 2.
Despite virtually returning to the depot, the outside vehicles are still required to start from the stop location when executing the new schedule after a change. So, these virtually returned vehicles have to first virtually move to their stop location in the new schedule. Therefore, the vehicles must serve some virtual paths in the new schedule to reach this stop location. The graph of the CARP instance is thus modified to include these virtual paths, which can be regarded as virtual tasks being optimised along with the normal tasks by a static CARP algorithm, as shown in Lines \ref{alg:vt-line2}-\ref{alg:vt-line4}. We also need all vehicles in the depot to have the same full capacities to be able to use static CARP algorithms. Therefore, we assign the previous demands that have been served by an outside vehicle to the corresponding virtual task, as shown in Line \ref{alg:vt-line7}. As a result, a DCARP instance is converted to a ‘static’ CARP instance, in which all vehicles are located at the depot with the same capacities.} \rev{It is worthy to mention that the ``virtual task'' idea has also been used for large scale CARP \cite{tang2016scalable}, which is totally different from our idea here. They used the virtual task to represent the grouped neighboring tasks such that the problem's dimensionality can be reduced.}

\begin{algorithm}
    \caption{Pseudocode of constructing virtual tasks} \label{vt-strategy}
    \KwIn{Task set $R = \{ t_1, t_2, ..., t_{N_t}\}$, \newline
    Stop locations of outside vehicles: $OV$, \newline
    Remaining capacity of outside vehicles: $RQ$. \newline
    $OV = \{ v_1, v_2, ..., v_{N_{ov}}  \}$,  
    $RQ = \{ q_1, q_2, ..., q_{N_{ov}} \}$ }
    \KwOut{The updated task set: $R$}
    \vspace{0.5em}
    \For{each outside vehicle $k$}
    {
        Construct an Arc with virtual task $vt_k$; \label{alg:vt-line2}\\ 
        Head: \rev{$head_{vt_k}$ = $v_0$}, where $v_0$ is the depot; \\
        Tail: \rev{$tail_{vt_k}$ = $v_k$}; \label{alg:vt-line4}\\
        Deadheading cost: \rev{$dc(vt_k)$ = \rev{$\infty$}}; \label{alg:vt-line5}\\
        Serving cost: \rev{$sc(vt_k)$ = $mdc(v_0, v_k)$}; \label{alg:vt-line6}\\ 
        Demand: \rev{$dm(vt_k)$} = $Q-q_k$, where $Q$ is the original capacity of vehicles; \label{alg:vt-line7} \\
        Add this virtual task into task set: $R = R \cup vt_k$.
    }
\end{algorithm}

A virtual task can also be \rev{interpreted} as a representation of an outside vehicle's previous serving \rev{status}, including the total cost, served demand and stop location before the occurrence of the dynamic \rev{events}. During the optimisation, the virtual tasks are regarded as arcs to be assigned to routes when being rescheduled. These arcs need to be served, so that some depot vehicles will actually correspond to the outside vehicles. Once a depot vehicle serves a virtual task, its remaining capacity will become the same as the remaining capacity of the corresponding outside vehicle, and so will its stop location. However, vehicles that are not serving these \sm{virtual} arcs should \rev{not be able} to traverse them, because these \sm{virtual} arcs are not actual physical paths that can be used by vehicles. This is achieved by assigning a deadheading cost (traversing cost) of $dc(vt) = \infty$ to these arcs, \rev{as shown in Line \ref{alg:vt-line5}}. Note that this infinite traversing cost is not included as part of the serving cost.

The serving cost of a virtual task \rev{should be} zero. This is because in reality, the outside vehicles are already in the stop locations and have already served some tasks. They should not \rev{incur} any extra cost to \rev{stay where they were}. However, in our strategy, a virtual task's serving cost is set as the \rev{minimal total deadheading cost} between the depot and the vehicle stop location \rev{(Line \ref{alg:vt-line6})} because some algorithms such as \rev{Path-Scanning use this cost as a denominator when deciding} which task to assign to the current route \cite{golden1983computational}. To avoid this cost being counted towards the total cost in the objective function, the additional cost will be subtracted from the actual total cost \rev{after} the optimisation \rev{using the virtual task strategy}. Besides, the demand is set as \sm{the} amount corresponding to the demand already served by the vehicle, i.e. $Q-q_k$, to avoid the total demand of tasks in its new route exceeding the vehicle's remaining capacity $q_k$ \rev{(Line \ref{alg:vt-line7})}.

\begin{figure}[h]
    \centering
    \includegraphics[width=\linewidth]{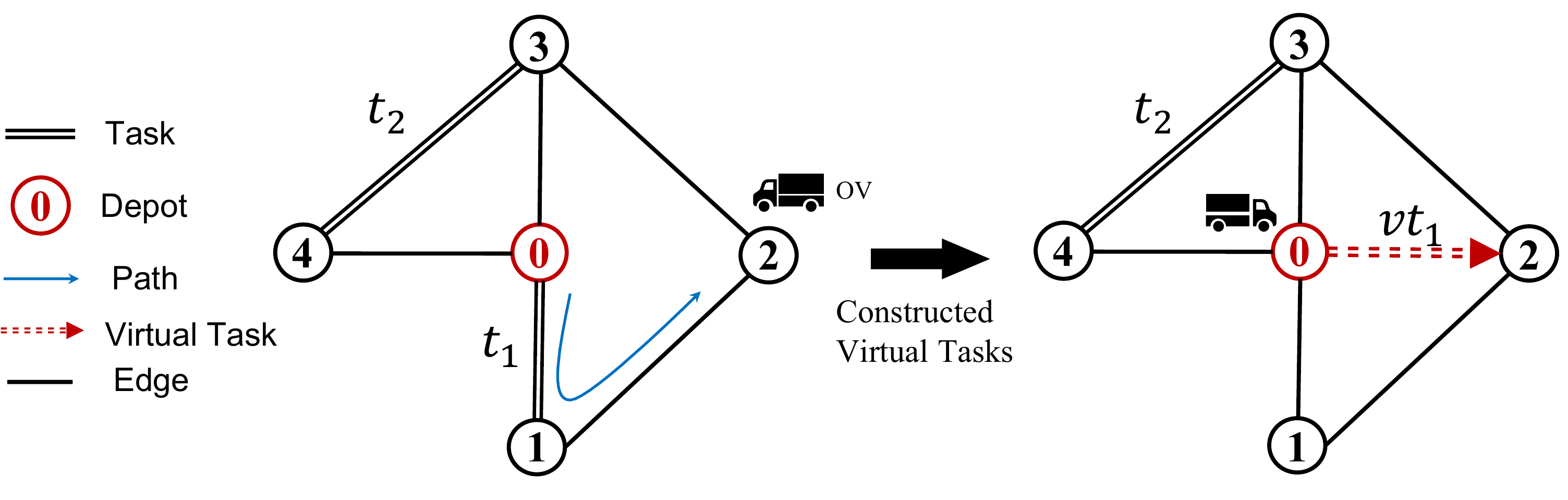}
    \caption{\rev{An example of constructing virtual tasks.}} \label{example-vt}
\end{figure}

In order to improve the understanding of how virtual tasks in DCARP instances are constructed, an example \rev{for tasks $t_1, t_2$} is provided in Figure \ref{example-vt}. \rev{One vehicle traverses from $v_0$ (depot) and serves the task $t_1$ (Figure \ref{example-vt} left).} When the dynamic change happens, a vehicle is located at $v_2$. This vehicle is virtually placed into the depot and a virtual task $vt_1$ is constructed between $v_0$ and $v_2$ \rev{(Figure \ref{example-vt} right)}. This virtual task will enable the vehicle to go back to its outside position to continue serving other tasks. \rev{The demand of $vt_1$, $dm(vt_1)$,  equals to $ Q-dm(t_1)$, such that when the vehicle virtually returns to its outside position, its remaining capacity will be the same as the remaining capacity at the rescheduling point. The deadheading cost of $vt_1$, $dc(vt_1)$ is set to $\infty$ to prevent other vehicles from traversing the virtual arc. 
This enables the outside vehicle to serve the virtual task. The serving cost of $vt_1$, $sc(vt_1)$, equals to $ mdc(v_0, v_2)$. This serving cost is incurred by the vehicle when it serves the virtual task, but is later on deducted from the objective function of the problem, so that the objective value with the virtual task strategy remains the same as the objective value without the virtual task strategy.} After that, the new DCARP instance with the remaining task $t_2$ and the virtual task $vt_1$ will be optimised using one of the algorithms which are available for static CARP, in which the task $t_1$ is removed because it was already served before the change that triggered the rescheduling. 

After \rev{applying} the virtual-task strategy, the route formulation in a DCARP solution $S = \{ r_1, r_2, ..., r_{N_{ov}}, ..., r_K \}$ becomes: \newline
\begin{equation*}
    \begin{aligned}
        &r_k = \rev{(} v_0, vt_k, t_{k, 2}, t_{k, 3}, ..., t_{k, l_k}, v_0\rev{)}, k = 1, 2, ..., N_{ov}. \\
        &r_k = \rev{(} v_0, t_{k, 1}, t_{k, 2}, ..., t_{k, l_k}, v_0\rev{)}, k = N_{ov}+1, ..., K.
    \end{aligned}
\end{equation*}

The new formulation of the optimisation objective \rev{and the constraints} for a DCARP instance is given in Eq. \ref{new-objective}:
\begin{equation}
    \begin{aligned}
        \mathbf{Min} &\quad TC(S) = \sum_{k=1}^{K} RC_{r_k} - \sum_{k=1}^{N_{ov}} mdc(v_0, v_k). \label{new-objective} \\
        \mathbf{s.t.} &\quad \sum_{k=1}^{K}l_k = N_t + N_{ov}. \\
        &\quad t_{k_1, i_1} \neq t_{k_2, i_2}, \rev{\mathrm{for\ all\ }} (k_1, i_1) \neq (k_2, i_2). \\
        &\quad \sum_{i=1}^{l_k}dm(t_{k, i}) \leq Q, \forall k = 1, ..., K. 
    \end{aligned}
\end{equation}
where for route $\{r_k | k = 1, 2, ..., N_{ov}\}$, $t_{k, 1} = vt_k$. \rev{The second term $\sum_{k=1}^{N_{ov}} mdc(v_0, v_k)$ is to balance out Line \ref{alg:vt-line6} in Algorithm \ref{vt-strategy} so that we do not count the serving costs of virtual tasks.}

\rev{The above} adjustments enable a new schedule for the converted `static' CARP instance to be obtained by directly using meta-heuristic algorithms for static CARP. An executable solution is obtained by removing the virtual tasks from the routes in the new schedule and assigning it to the corresponding outside vehicles.
\rev{The virtual task is better to be the first task in a vehicle's route. If a given virtual task is not the first task of a service route found by the static CARP algorithm, the tasks before this virtual task will be assigned to a new vehicle starting from the depot and the following tasks will be served by the corresponding outside vehicle. In such a case, we split a route that contains virtual tasks in the middle into multiple routes with one route served by a vehicle from the depot and others served by the corresponding outside vehicle. }
\rev{In this way, the total cost of the CARP solution will not be influenced by splitting because the head node of the virtual task is also the depot.} \rev{An} example \rev{(without relations to Figure \ref{example-vt})} of converting an obtained new solution to an executable service plan is provided below: 
\begin{equation*}
    \begin{array}{c}
        \rev{( v_0, vt_1, t_2, v_0) , (v_0, t_{3}, t_{4}, vt_{2}, t_{5}, v_0 )}\\
        \downarrow \\
        \rev{( v_0, vt_1, t_2, v_0) , (v_0, t_{3}, t_{4}, v_0) , (v_0, vt_{2}, t_{5}, v_0)}\\
        \downarrow \\
        \rev{( v_1, t_2, v_0) , (v_0, t_{3}, t_{4}, v_0) , (v_2, t_{5}, v_0)}.
    \end{array}
    \end{equation*}
\rev{In the above example, the solution (top) contains two routes and two outside vehicles (i.e. two virtual tasks). In the second route, the virtual task located in the middle of the task sequence. If the first task of a route is a virtual task, the remaining tasks of the route are served by the outside vehicle. Otherwise, all tasks of the route are served by a new empty vehicle. Therefore, after conversion (middle), tasks $t_3, t_4$ are assigned to a new vehicle starting from the depot, but task $t_5$ should actually be served by the outside vehicle. The final executable routes (bottom), are obtained by removing the virtual tasks and starting from the corresponding stop locations (The $v_1$ and $v_2$ are the stop locations of outside vehicles corresponding to $vt_1$ and $vt_2$).}

\subsection{Proposed Framework based on Virtual-Task Strategy}

There are generally two commonly used strategies in dynamic optimisation. \rev{One is to} re-start the optimisation, which \rev{can also include} some additional diversity enhancing techniques \cite{mavrovouniotis2017survey}. The other is to migrate some good solutions \rev{for} the old environment to the new environment and initialise the starting individuals with them \cite{nguyen2012survey,mavrovouniotis2013adapting}. 

These two strategies \rev{could} also be used in \rev{our} optimisation framework. The restart strategy is \rev{straightforward to apply after a change has occurred}. However, the current strategies for \rev{re-using good solutions for the old environment may not be} suitable for \rev{our} DCARP scenarios because \rev{the} dynamic events \rev{may} influence the problem's dimension and the previous solution's feasibility in a DCARP scenario. For example, the served tasks and added tasks change the total number of tasks, and the potential road closure event \rev{makes previous solutions} infeasible in the new DCARP instance.

Therefore, we propose a new \rev{knowledge (especially sequence)} transfer strategy for the DCARP scenario \sm{(Algorithm \ref{inheriting-strategy})}, which can benefit from previous DCARP instance's best solutions. Given an instance $I_m$, all scheduled routes in the best solution $S_{m-1}$ of the previous instance $I_{m-1}$ are concatenated to construct an ordered list of tasks $P_{m-1}$. All tasks that have been served are then removed from the list in $P_{m-1}$, and the remaining tasks keep their orders. After that, the newly added tasks are inserted into $P_{m-1}$ greedily, \sm{i.e.,} they are inserted into the positions with \sm{the} smallest increased cost. Finally, a corresponding transferred solution is generated by using the split scheme to convert \rev{an} ordered list to an explicit-route solution. The newly transferred solution is used as one of the initial solutions when optimising the new DCARP instance \rev{using} the population-based algorithm.

\begin{algorithm}[!htpb]
    \caption{Pseudocode of \rev{knowledge (sequence)} transfer strategy} \label{inheriting-strategy}
    \KwIn{Previous best solution $S_{m-1}$ }
    \KwOut{A newly transferred solution $S_{m}$}
    Convert $S_{m-1}$ into an ordered list of tasks $P_{m-1}$; \\
    Remove all served tasks from $P_{m-1}$; \\
    $P_m = P_{m-1}$, $l_P = |P_{m}|$; \\
    The newly added tasks set $NT = \{nt_1, nt_2, ..., nt_{|NT|} \}$; \\
    \For{each $nt_i \in NT$}
    {
        \For {each position in $P_{m}$}
        {
            Calculate the increased cost after insertion;
        }
        Obtain the position $p$ with the smallest increased cost; \\
        Insertion: $P_m = [t_1, ..., t_p, nt_i, t_{p+1}, ..., t_{l_P}]$; \\ 
        $l_P = l_P + 1$; \\
        $NT = NT \setminus \{nt_i\}$. \\
    }

    Use split scheme in $P_m$. \\
    
\end{algorithm}

On the basis of the virtual-task strategy and two initialisation strategies, i.e. the restart strategy and the sequence transfer strategy, \rev{we propose a generalised optimisation framework with virtual tasks (GOFVT) to generalise} static CARP algorithms to dynamic scenarios. \rev{Our framework comprises four main steps}:
\begin{enumerate}
    \item Construct virtual tasks;
    \item Apply the restart strategy (randomly generate initial solutions) or the sequence transfer strategy (generate one transferred solution for individual-based algorithms and additionally generated random solutions for population-based algorithms);
    \item Apply the meta-heuristic algorithm to optimize the converted ``static'' CARP instance;
    \item Convert the obtained solution with virtual tasks to an executable solution \rev{without virtual tasks}.
\end{enumerate}
For a DCARP instance, the framework will first construct the virtual tasks to convert the dynamic instance to a `static' instance. Then, one of the two initialisation strategies explained above can be adopted to assist the optimisation for the DCARP instance. In the computational studies in Section \ref{experiment}, we will compare the effectiveness of these two strategies. Then, meta-heuristic algorithms with the initialisation strategy are applied to optimise the DCARP instance \rev{which is} a `static' instance with virtual tasks. Finally, the solution obtained for the `static' instance is converted into an executable solution, in which the routes with virtual tasks are assigned to the corresponding outside vehicles and the routes without virtual tasks are assigned to vehicles \sm{located at} the depot. 

\section{Computational Studies}\label{experiment}
In order to evaluate the efficiency of our proposed framework (GOFVT), three sets of experimental studies have been conducted by embedding a selection of meta-heuristic algorithms in the GOFVT in this section. \rev{After providing the experimental set-up (Section V-A), in} the first experiment, the virtual-task strategy is compared with a simple rescheduling strategy (\rev{Section \ref{experiment1}}). After that, the virtual-task strategy's efficiency is investigated by comparing it with an existing algorithm for DCARP \rev{in the second experiment} (\rev{Section \ref{experiment2}}). Finally, \rev{in the last experiment}, GOFVT is combined with several classical meta-heuristic algorithms originally designed for static CARP, and its performance is analysed by running the newly generated algorithms in DCARP scenarios (\rev{Section \ref{experiment3}}).

\subsection{Experimental \sm{Settings}}
All experiments are conducted on a series of DCARP instances or scenarios generated by the simulation system presented in Section \ref{simulation-system}, based on a static CARP benchmark, namely the \textit{egl} set \cite{li1996interactive}. The \textit{egl} set contains 24 CARP instances. In each experiment, the DCARP instances or scenarios are generated independently from static CARP instances. 
For our first and second set of experiments, 3 DCARP instances are generated \rev{for each static CARP instance}. These DCARP instances are not used to compose a DCARP scenario, as the algorithms just optimise the current instance and \rev{do not use any knowledge transfer} in these two experiments.  
For the third \rev{experiment}, one DCARP scenario including 5 DCARP instances is generated based on each CARP instance. \rev{For a fair comparison, each DCARP instance is generated from the best solution among all obtained solutions by all compared algorithms in the previous instance.}
When the simulator executes \rev{the selected} solution according to the deployment policy, the time for serving all tasks will be calculated first and the simulator will \rev{uniformly} randomly select a stop point \rev{(for dynamic events)} within the longest time. \rev{As we only requires a set of DCARP instances to test the effectiveness of proposed strategies and framework, we have arbitrarily chosen parameters according to real world situations. For example, a road is more likely to become congested than being closed, hence we set $p_{road} = 0.1$. In our experiments, the parameters of the simulator are chosen as $p_{event}=0.5$, $p_{road}=0.1$, $p_{bdrr}=0.5$, $p_{crr}=0.3$, $p_{crbb}=0.6$, $p_{cid}=0.35$, $p_{add}=0.35$.} \rev{In the future, we will carry out a more comprehensive study of the characteristics of simulator in relation to its parameter values.} Eight different types of dynamic changes (Table \ref{dynamic-change}) are simulated in \rev{our} simulator excluding the case of ``vehicles broken down'' because its formulation is the same as the event of added tasks. 
\rev{As an illustrative example, we investigate the influence of different scenarios on the optimisation algorithms’ performances. These examples show that the scenario with newly added tasks has a significant influence on the optimisation algorithm's performance\footnote{More details are in Section IV and Table VI of the supplementary material.}.}

Because a key optimisation requirement when dynamic changes happen in the real world is to obtain a new solution quickly, we limit the maximum optimisation time to 60s for the small problems (E1 $\sim$ E4) and 180s for larger maps (S1 $\sim$ S4) in \textit{egl} for all algorithms. \rev{All programs are implemented in C language and run using a PC with an Intel Core i7-8700 3.2GHZ.} The source code of  our experiments has been available on github\footnote{\rev{https://github.com/HawkTom/Dynamic-CARP}}. 

\subsection{Is It Necessary to Reschedule for DCARP?}\label{experiment1}
Although \rev{many} algorithms have been proposed to solve DCARP in the literature, a simple baseline strategy, named the return-first strategy, has been ignored. The return-first strategy schedules all outside vehicles back to the depot first in order to convert a DCARP instance to a static one, and then reschedules all vehicles \rev{for the} new static instance after all vehicles are located at the depot. If the return-first strategy is efficient enough, the direct optimisation of a DCARP instance would not be necessary any more. However, \rev{since this has not been shown in the literature so far, in this subsection}, we use the proposed virtual-task strategy to solve DCARP and compare it with the return-first strategy to show the importance of optimising DCARP instances directly instead of ignoring the outside vehicles and assigning new vehicles to all remaining tasks.

\begin{table*}[!htpb]
    \renewcommand{\arraystretch}{1.2}
    \centering
    \caption{Results of virtual-task strategy (VT) and return-first strategy (RF) on DCARP instances with different settings of remaining capacities from the \textit{egl} dataset. The value in each cell represents ``Mean $\pm$ Std'' over 25 independent runs and the bold ones denote the better result on the DCARP instance based on \sm{the} Wilcoxon signed-rank test with a significance level of 0.05. The penultimate row summaries the number of win-draw-lose of RF strategy versus VT strategy and the last row provides the p-values of \sm{the} Wilcoxon signed-rank test with a significance level of 0.05 on instances with the same \sm{settings} of all maps.}\label{e1}

    \begin{adjustbox}{width = 0.8\textwidth, keepaspectratio}
        \begin{tabular}{@{}ccccccc@{}}
            \toprule
            \multirow{2}{*}{Static Map} & \multicolumn{2}{c}{Instance 1 ($q\in[0, 0.33Q]$)}         & \multicolumn{2}{c}{Instance 2 ($q\in[0.34Q, 0.66Q]$)} & \multicolumn{2}{c}{Instance 3 ($q\in[0.67Q, Q]$)} \\ \cmidrule(l){2-7} 
            \multicolumn{1}{l}{}                     & RF                             & VT                             & RF            & VT                             & RF            & VT                             \\ \midrule
            E1-A & 8386$\pm$22 & \bf{\highlight{8095$\pm$31}} & 10397$\pm$34 & \bf{\highlight{9925$\pm$22}} & 8630$\pm$8 & \bf{\highlight{8566$\pm$10}} \\ 
            E1-B & \bf{\highlight{9830$\pm$29}} & \bf{\highlight{9839$\pm$39}} & 9718$\pm$22 & \bf{\highlight{8437$\pm$28}} & 9675$\pm$16 & \bf{\highlight{9208$\pm$29}} \\ 
            E1-C & \bf{\highlight{12181$\pm$52}} & \bf{\highlight{12109$\pm$52}} & 14726$\pm$31 & \bf{\highlight{14106$\pm$16}} & 11286$\pm$11 & \bf{\highlight{10565$\pm$25}} \\ 
            E2-A & \bf{\highlight{10255$\pm$6}} & 10480$\pm$6 & 13740$\pm$27 & \bf{\highlight{12278$\pm$29}} & 11692$\pm$51 & \bf{\highlight{10231$\pm$34}} \\ 
            E2-B & \bf{\highlight{13162$\pm$15}} & 13317$\pm$23 & 16435$\pm$29 & \bf{\highlight{15207$\pm$18}} & 15667$\pm$35 & \bf{\highlight{14321$\pm$55}} \\ 
            E2-C & \bf{\highlight{17425$\pm$41}} & 17566$\pm$35 & 17444$\pm$32 & \bf{\highlight{16881$\pm$25}} & 14186$\pm$16 & \bf{\highlight{13871$\pm$46}} \\ 
            E3-A & 11343$\pm$42 & \bf{\highlight{11167$\pm$17}} & 12877$\pm$21 & \bf{\highlight{12030$\pm$26}} & 10369$\pm$19 & \bf{\highlight{8869$\pm$12}} \\ 
            E3-B & \bf{\highlight{15517$\pm$84}} & 15776$\pm$31 & 15150$\pm$31 & \bf{\highlight{12797$\pm$68}} & 20743$\pm$59 & \bf{\highlight{20081$\pm$39}} \\ 
            E3-C & \bf{\highlight{20549$\pm$69}} & 20880$\pm$63 & 20012$\pm$144 & \bf{\highlight{18078$\pm$88}} & 27521$\pm$81 & \bf{\highlight{25289$\pm$82}} \\ 
            E4-A & \bf{\highlight{11216$\pm$31}} & 11387$\pm$27 & 12489$\pm$26 & \bf{\highlight{12152$\pm$32}} & 16213$\pm$29 & \bf{\highlight{14858$\pm$135}} \\ 
            E4-B & 15522$\pm$40 & \bf{\highlight{15242$\pm$62}} & 14933$\pm$65 & \bf{\highlight{13791$\pm$63}} & 17477$\pm$71 & \bf{\highlight{15869$\pm$29}} \\ 
            E4-C & 20928$\pm$63 & \bf{\highlight{20691$\pm$41}} & 21998$\pm$28 & \bf{\highlight{20174$\pm$86}} & 17869$\pm$79 & \bf{\highlight{15798$\pm$59}} \\ 
            S1-A & 16097$\pm$49 & \bf{\highlight{16012$\pm$40}} & 15643$\pm$31 & \bf{\highlight{14917$\pm$40}} & 13684$\pm$53 & \bf{\highlight{13159$\pm$18}} \\ 
            S1-B & \bf{\highlight{20462$\pm$69}} & 21167$\pm$58 & 16184$\pm$35 & \bf{\highlight{15697$\pm$35}} & 15151$\pm$19 & \bf{\highlight{13428$\pm$82}} \\ 
            S1-C & \bf{\highlight{27944$\pm$71}} & 28624$\pm$116 & 21040$\pm$36 & \bf{\highlight{18616$\pm$87}} & 26470$\pm$103 & \bf{\highlight{25789$\pm$76}} \\ 
            S2-A & 19610$\pm$85 & \bf{\highlight{19479$\pm$107}} & 22314$\pm$40 & \bf{\highlight{20016$\pm$48}} & 22499$\pm$67 & \bf{\highlight{19212$\pm$82}} \\ 
            S2-B & \bf{\highlight{27970$\pm$75}} & 28408$\pm$59 & 26334$\pm$130 & \bf{\highlight{25911$\pm$118}} & 25723$\pm$67 & \bf{\highlight{25024$\pm$94}} \\ 
            S2-C & \bf{\highlight{31859$\pm$108}} & 32101$\pm$125 & 40170$\pm$132 & \bf{\highlight{38575$\pm$186}} & 35053$\pm$143 & \bf{\highlight{31815$\pm$133}} \\ 
            S3-A & 19489$\pm$64 & \bf{\highlight{19211$\pm$107}} & 23630$\pm$118 & \bf{\highlight{21329$\pm$98}} & 28237$\pm$76 & \bf{\highlight{24494$\pm$71}} \\ 
            S3-B & \bf{\highlight{27518$\pm$72}} & \bf{\highlight{27521$\pm$98}} & 25651$\pm$64 & \bf{\highlight{24311$\pm$93}} & 32286$\pm$50 & \bf{\highlight{30646$\pm$96}} \\ 
            S3-C & 32468$\pm$93 & \bf{\highlight{32049$\pm$107}} & 35435$\pm$67 & \bf{\highlight{33927$\pm$68}} & 43151$\pm$98 & \bf{\highlight{40801$\pm$97}} \\ 
            S4-A & \bf{\highlight{27256$\pm$87}} & 28082$\pm$142 & 23780$\pm$74 & \bf{\highlight{22788$\pm$95}} & 29895$\pm$102 & \bf{\highlight{27576$\pm$105}} \\ 
            S4-B & 34872$\pm$122 & \bf{\highlight{34709$\pm$149}} & 28283$\pm$121 & \bf{\highlight{28023$\pm$93}} & 31770$\pm$87 & \bf{\highlight{30652$\pm$125}} \\ 
            S4-C & \bf{\highlight{43746$\pm$168}} & 44165$\pm$157 & 36808$\pm$117 & \bf{\highlight{36706$\pm$123}} & 44006$\pm$145 & \bf{\highlight{42827$\pm$178}} \\ \midrule

            $\#$of `w-d-l'    &  \multicolumn{2}{c}{12-3-9}                   &  \multicolumn{2}{c}{0-0-24} & \multicolumn{2}{c}{0-0-24 }
            \\
            p-value    & \multicolumn{2}{c}{0.22}                 &\multicolumn{2}{c}{1.82e-5}&\multicolumn{2}{c}{1.82e-5}
            \\ \bottomrule
            \end{tabular}    
    \end{adjustbox}

\end{table*}

\begin{table}[!htpb]
    \renewcommand{\arraystretch}{1.2}
    \centering
    \setlength{\tabcolsep}{10pt}
    \caption{Results of MASDC and VT-MASDC on DCARP instances from the \textit{egl} dataset. The value in each cell represents ``Mean $\pm$ Std'' over 25 independent runs and the bold ones denote the better result on the DCARP instance based on \sm{the} Wilcoxon signed-rank test with a significance level of 0.05. The last row summaries the number of win-draw-lose of MASDC versus VT-MASDC.}\label{e2}
        \begin{tabular}{cccccc}
            \toprule
    Map                   & Ins.Index & MASDC          & VT-MASDC                        \\ \midrule
\multirow{3}{*}{S4-A} & 1   & 58456$\pm$1621 & \bf{\highlight{51512$\pm$3298}} \\
                      & 2   & 55394$\pm$1746 & \bf{\highlight{49422$\pm$3517}} \\
                      & 3   & 64795$\pm$1546 & \bf{\highlight{58676$\pm$3517}} \\
\multirow{3}{*}{S4-B} & 1   & 64343$\pm$1886 & \bf{\highlight{57017$\pm$3622}} \\
                      & 2   & 65027$\pm$1712 & \bf{\highlight{57413$\pm$3033}} \\
                      & 3   & 60143$\pm$1612 & \bf{\highlight{53342$\pm$3323}} \\
\multirow{3}{*}{S4-C} & 1   & 69224$\pm$1475 & \bf{\highlight{62760$\pm$2929}} \\
                      & 2   & 68070$\pm$1424 & \bf{\highlight{62304$\pm$2465}} \\
                      & 3   & 78134$\pm$1893 & \bf{\highlight{69348$\pm$3451}} \\ 
                      \midrule
                      \multicolumn{3}{l}{Statistical results over 72 instances below} &  \\
                      \midrule 
                      \multicolumn{2}{l}{$\#$ of win-draw-lose} & & 0-0-72 \\
                      \multicolumn{2}{c}{p-value} & & 1.67e-13 \\
                      \bottomrule
        \end{tabular}
    \begin{tablenotes} 
		\item \rev{The data of E1-A $\sim$ S3-C are omitted in the table due to the page limitation. The complete data are provided in the supplementary material. }
    \end{tablenotes}

\end{table}

In our experiment, the simulation system generates three different DCARP instances for each test map \rev{(i.e., a static instance in the \textit{egl} benchmark set)} with different sets of remaining capacities. \rev{As this experiment aims to demonstrate the necessity and efficiency of directly optimising DCARP instances, the setting of remaining capacities is divided into three intervals, i.e., $[0, 0.33Q]$,  $[0.34Q, 0.66Q]$ and $[0.67Q, Q]$.} Then, an optimisation algorithm, \sm{Memetic Algorithm with Extended Neighborhood Search (MAENS)} \cite{tang2009memetic}, assisted with the return-first strategy and virtual-task strategy \rev{are} applied to optimise each DCARP instance, respectively. \rev{Two algorithm instantiations using MAENS follow the same setting during the optimisation that the return-first strategy and virtual-task strategy is the only difference between them.} The comparison results in terms of mean and standard deviation over 25 independent runs (mean$\pm$std), of the return-first (RF) strategy and virtual-task (VT) strategy on DCARP instances with different remaining capacities are presented in Table \ref{e1}. The bold values with grey background for each DCARP instance are the better results between return-first strategy and virtual-task strategy based on the Wilcoxon signed-rank test with a significance level of 0.05. The second last row of Table \ref{e1} summarises the number of win-draw-lose of \rev{the RF versus VT strategies}. We have calculated the Wilcoxon signed-rank test with a significance level of 0.05 for the mean total cost of RF and VT strategies on the instances with the same range of remaining capacities, and the p-values are listed in the last row of Table \ref{e1}.

Table \ref{e1} shows that the \rev{RF and VT strategies} are significantly different \sm{for the instances where} the remaining capacities are in the range of $[0.34Q, Q]$ (instances 2 and 3 in Table \ref{e1}), with the VT strategy outperforming the RF strategy on all DCARP instances. In contrast, \sm{for} the scenarios with remaining capacities in $[0,0.33Q]$, there are 12 out of 24 DCARP instances where the RF strategy outperforms the VT strategy. When comparing the results using \sm{the} Wilcoxon test across maps, \sm{we confirm} that none of these strategies is a \rev{consistent} winner when analysed across maps \sm{where} the remaining capacities are smaller than $0.33Q$. This is understandable because when the vehicles are mostly \rev{fully loaded}, i.e., when the remaining capacities are smaller than 0.33Q, there is limited space for serving more tasks no matter what strategy is used.

\rev{In order to avoid the conclusion being biased by the employed meta-heuristic algorithm, we have employed another meta-heuristic algorithm, i.e., ILMA \cite{mei2009improved}, to execute the same experiment. Due to the page limitation, we put the results into Table V of the supplementary material. The statistical analysis has confirmed the same conclusion as the experiments employing MAENS.}

Overall, we can conclude that it is necessary and much more effective to optimise the DCARP instance directly rather than using the RF strategy when outside vehicles have enough remaining capacities.

\begin{figure}[!htpb]
    \centering
    \includegraphics[width=0.65\linewidth]{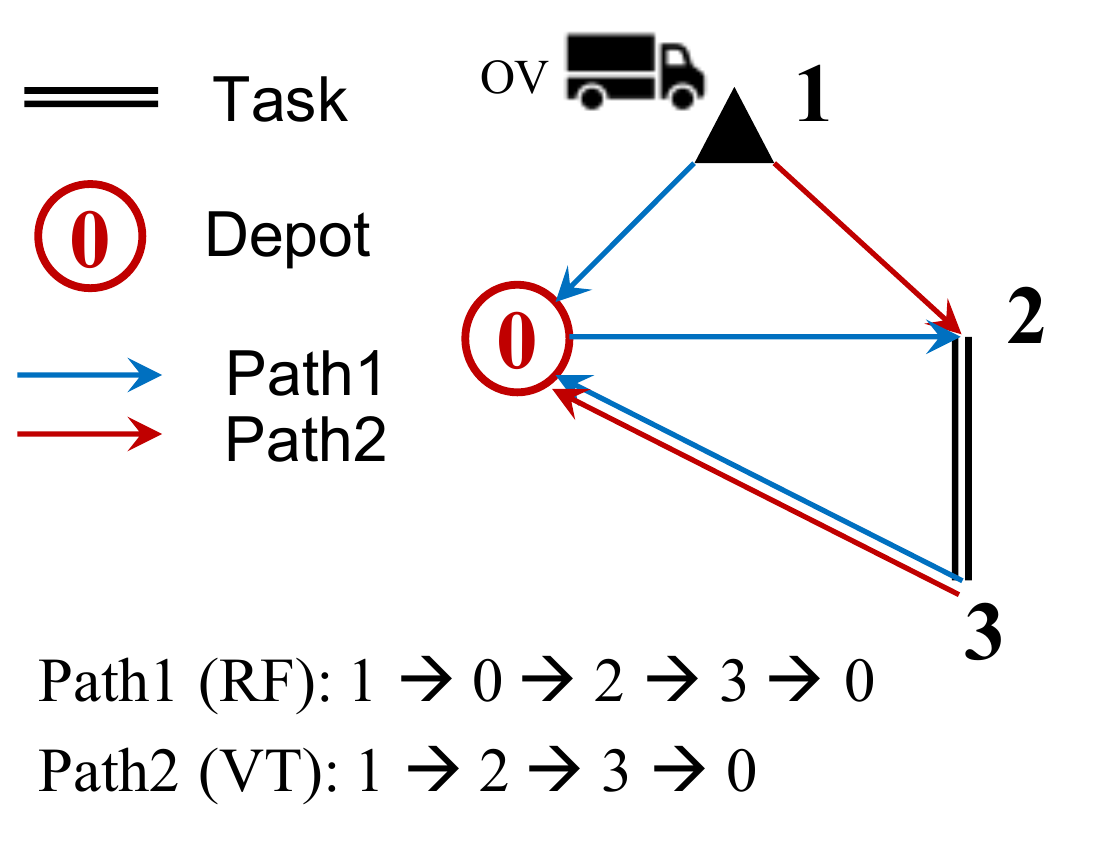}
    \caption{An example of demonstrating why the RF strategy is not efficient enough when outside vehicles have enough remaining capacities.} \label{example-rf}
\end{figure}

The reason why the RF strategy is not always helpful when outside vehicles have enough remaining capacities can be explained using a simple example \rev{in Figure} \ref{example-rf}, where an outside vehicle stops at vertex 1. If its remaining capacity is \sm{sufficient} to serve task $t_{23}$, it can directly traverse from vertex 1 to vertex 2, presented as `Path 2' in Figure \ref{example-rf}, and the final total cost will be $d_{vt} = d_{12} + d_{23} + d_{30}$. But if we apply the RF strategy, the total cost will change to $d_{rf} = d_{10} + d_{02} + d_{23} + d_{30}$, presented as `Path 1' in Figure \ref{example-rf}. It is obvious that $d_{rf} \geq d_{vt}$ because $d_{10} + d_{02} \geq d_{12}$ \rev{according to triangle inequality}. The RF strategy increases the final cost because vehicles take a detour in such cases.

\subsection{Analysis of the Effects of \sm{the} Virtual-Task Strategy}\label{experiment2}

Memetic \rev{algorithm with new split scheme (MASDC) \cite{liu2014memetic} is the only meta-heuristic algorithm for DCARP} in the literature that considers a general DCARP scenario including several dynamic events, such as road closure and added tasks. It comprises a distance-based split scheme to assist the DCARP solution being used in the crossover and local search. Our virtual-task strategy can convert a DCARP instance to a `static' CARP instance so that the operator used in the static CARP can be used in the DCARP instance directly. In this subsection, we analyse the effects of \rev{our} VT strategy by embedding it to MASDC, referred to as VT-MASDC, and comparing it to the original MASDC. The advantage of embedding \rev{our} strategy into MASDC is that this enables us to isolate and analyse the effect of the virtual tasks compared to a state-of-the-art DCARP algorithm.  
\rev{In particular, all components in MASDC and VT-MASDC are the same except for the use of virtual tasks and the distance-based split scheme. The latter needs to be replaced by Ulusoy's split scheme in VT-MASDC because the distance-based split scheme is specifically designed for DCARP instances with outside vehicles at different stop locations. When we apply the virtual task strategy, the DCARP instance is converted to the `static' instance where all vehicles are located at the depot. Therefore, the distance-based split scheme is not suitable anymore and the Ulusoy's split scheme is used instead. }
The use of virtual tasks is thus inherently linked to this split scheme, and any advantages provided by the virtual tasks are also linked to the fact that they enable this split scheme to be adopted.

In our experiment, we generate three independent DCARP instances for \rev{each map in the \textit{egl} benchmark set}, ensuring that outside vehicles has enough remaining capacities\sm{, i.e. $q \geq 0.5Q$,} in all generated instances. \sm{However, we} need to generate DCARP instances rather than DCARP scenarios because the aim of DCARP is to minimise the total cost for each DCARP instance separately \sm{(Eq. \ref{objective2})}. The optimisation results are presented in Table \ref{e2}, in which the values in each cell represent the mean and standard deviation over 25 independent runs (mean$\pm$std). For each DCARP instance, the better result, based on the Wilcoxon signed-rank test with a significance level of 0.05, is highlighted \sm{using a} bold font and grey background in Table \ref{e2}. The summary of win-draw-lose of MASDC versus VT-MASDC, which is presented in the last row, shows that VT-MASDC outperforms MASDC on all generated DCARP instances. The Wilcoxon signed-rank test with a significance level of 0.05 was conducted for the average total cost of VT-MASDC and MASDC in all instances, and the p-value was 1.67e-13. Overall, we can conclude that VT-MASDC performs much better than the original MASDC.

MASDC uses a distance-based split scheme (DSS) to evaluate DCARP solution's fitness because the split scheme designed for static CARP is not suitable for DCARP \cite{tong2020towards}. The DSS operator randomly splits the sequence of tasks to an executable CARP solution with explicit routes, and the random splitting process is repeated three times to obtain a better schedule. After embedding with the VT strategy, the DSS operator is replaced by Ulusoy's split \cite{ulusoy1985fleet} as explained in the beginning of this \rev{subsection}, and the evaluation of a sequence of tasks only requires \sm{to apply the Ulusoy's split once}. As a result, the randomness brought by \sm{the} DSS's split scheme is removed.

\begin{figure}[!htpb]
    \centering
    \includegraphics[width=0.7\linewidth]{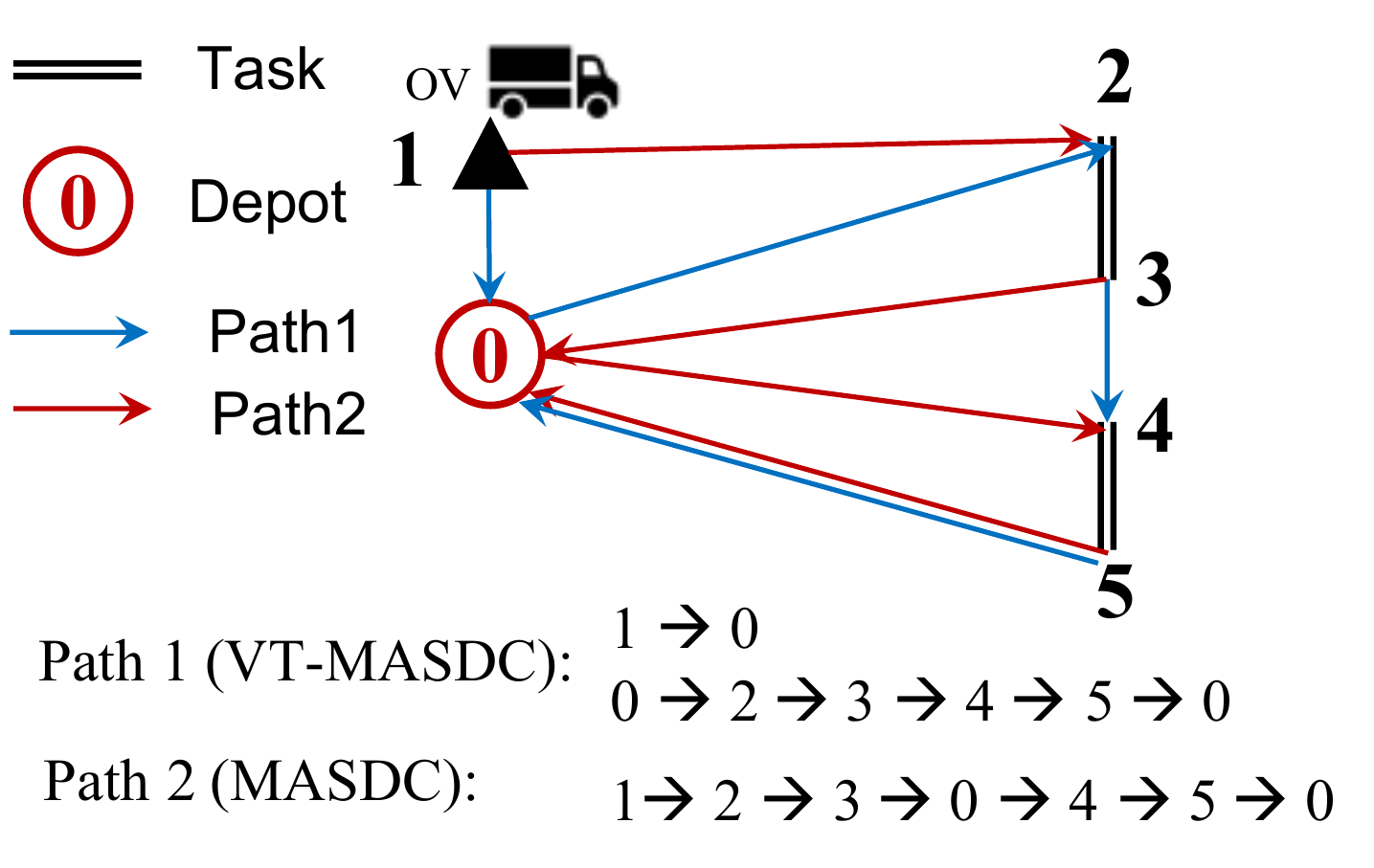}
    \caption{An example of demonstrating the advantages of considering new vehicles.} \label{example-newv}
\end{figure}

Moreover, the DSS operator never considers new vehicles starting from the depot during the optimisation, whereas our VT strategy enables both outside and new vehicles to be used. We provide an example in Figure \ref{example-newv} to show the advantages of considering new vehicles during the optimisation \rev{of} DCARP. An outside vehicle is located at vertex 1, and its remaining capacity can only serve task $t_{23}$. We assume that $dd_{10} + dd_{02} \approx dd_{12}$ and $dd_{30} + dd_{04} \gg dd_{34} $ and the total cost of `Path1' and `Path2' can be calculated as:
\begin{equation*}
\begin{array}{c}
    TC_{1} = \underline{\rev{dd_{10}} + dd_{02}} + sc_{23} + \underline{dd_{34}} + sc_{45} + dd_{50}, \\
    TC_{2} = \underline{dd_{12}} + sc_{23} + \underline{dd_{30} + dd_{04}} + sc_{45} + dd_{50}. \\
\end{array}
\end{equation*}
For a sequence of tasks $[0, t_{23}, t_{45}, 0]$, if applied with the DSS operator, the only obtained path will be `Path 2' as shown in Figure \ref{example-newv}. In contrast, if we use the VT strategy, a better path, i.e. `Path 1' in Figure \ref{example-newv}, can be obtained, which \sm{avoids} traversing the longer returning path.

\begin{table*}[!htpb] 
    \centering
    \caption{Results of VT-RTS, VTtr-RTS, VT-ILMA, VTtr-ILMA, VT-MAENS, VTtr-MAENS, VT-MASDC and VTtr-MASDC on the \textit{egl} dataset. The values in each cell \sm{represent} the ``MEAN$\pm$STD'' with the average ranking (in the brackets) w.r.t the average total cost over 25 independent runs. The bold values are the better \sm{results} under the Wilcoxon signed-rank test with a significance level of 0.05 between restart and inheriting strategies in an algorithm for a DCARP instance. The last three rows summarise the number of win-draw-lose of restart strategy versus inheriting strategy in each algorithm and the p-values of \sm{the} Wilcoxon signed-rank test with a significance level of 0.05 on all instances. }\label{e3}
    \begin{adjustbox}{max totalheight=\textheight, keepaspectratio}
        \begin{adjustbox}{width=0.95\textwidth, keepaspectratio}
        \begin{tabular}{@{}cccccccccc@{}}
            \toprule
            Map                                            & Ins.Index                          & VT-RTS                               & VTtr-RTS                             & VT-ILMA                              & VTtr-ILMA                            & VT-MAENS                            & VTtr-MAENS                          & VT-MASDC         &VTtr-MASDC                        \\ \midrule
            \multirow{5}{*}{S3-A} & 1 & \bf{\textit{22989$\pm$156(3.6)}} & \bf{\textit{22971$\pm$190(3.4)}} & \bf{\textit{24100$\pm$307(5.5)}} & \bf{\textit{24073$\pm$289(5.5)}} & \bf{\textit{22411$\pm$61(1.5)}} & \bf{\textit{22401$\pm$70(1.5)}} & \bf{\textit{52327$\pm$3849(7.5)}} & \bf{\textit{52327$\pm$3849(7.5)}} \\ 
            & 2 & \bf{\textit{25249$\pm$348(2.5)}} & 26012$\pm$192(4.0) & \bf{\textit{27134$\pm$338(5.3)}} & \bf{\textit{27234$\pm$405(5.6)}} & \bf{\textit{25013$\pm$105(1.8)}} & \bf{\textit{25017$\pm$103(1.8)}} & \bf{\textit{57870$\pm$3931(7.5)}} & \bf{\textit{57870$\pm$3931(7.5)}} \\ 
            & 3 & \bf{\textit{25907$\pm$350(3.3)}} & 26117$\pm$160(3.7) & \bf{\textit{27011$\pm$250(5.5)}} & \bf{\textit{27042$\pm$320(5.5)}} & \bf{\textit{25285$\pm$145(1.4)}} & \bf{\textit{25349$\pm$113(1.6)}} & \bf{\textit{59652$\pm$3792(7.5)}} & \bf{\textit{59652$\pm$3792(7.5)}} \\ 
            & 4 & 41928$\pm$177(6) & \bf{\textit{33347$\pm$287(3)}} & \bf{\textit{34800$\pm$398(4.5)}} & \bf{\textit{34840$\pm$420(4.5)}} & \bf{\textit{32374$\pm$119(1.4)}} & \bf{\textit{32423$\pm$100(1.6)}} & \bf{\textit{72252$\pm$4788(7.5)}} & \bf{\textit{72250$\pm$4790(7.5)}} \\ 
            & 5 & 31319$\pm$4(6) & \bf{\textit{24329$\pm$316(2.7)}} & \bf{\textit{25614$\pm$322(4.6)}} & \bf{\textit{25632$\pm$334(4.4)}} & \bf{\textit{23989$\pm$74(1.7)}} & \bf{\textit{24006$\pm$85(1.6)}} & \bf{\textit{62362$\pm$4014(7.5)}} & \bf{\textit{62362$\pm$4014(7.5)}} \\ 
            \multirow{5}{*}{S3-B} & 1 & 30945$\pm$70(6) & \bf{\textit{29429$\pm$141(5)}} & \bf{\textit{26486$\pm$282(3.4)}} & \bf{\textit{26493$\pm$230(3.6)}} & \bf{\textit{25161$\pm$123(1.5)}} & \bf{\textit{25165$\pm$106(1.5)}} & \bf{\textit{46491$\pm$3247(7.5)}} & \bf{\textit{46491$\pm$3247(7.5)}} \\ 
            & 2 & \bf{\textit{28494$\pm$1307(3.5)}} & 29198$\pm$1867(4.1) & \bf{\textit{29318$\pm$356(5.2)}} & \bf{\textit{29374$\pm$347(5.2)}} & 27445$\pm$121(1.8) & \bf{\textit{27356$\pm$140(1.2)}} & \bf{\textit{55642$\pm$2518(7.5)}} & \bf{\textit{55642$\pm$2518(7.5)}} \\ 
            & 3 & \bf{\textit{36665$\pm$2093(5.3)}} & \bf{\textit{35529$\pm$3448(3.8)}} & \bf{\textit{34378$\pm$330(4.2)}} & \bf{\textit{34515$\pm$347(4.7)}} & \bf{\textit{32584$\pm$78(1.5)}} & \bf{\textit{32613$\pm$112(1.5)}} & \bf{\textit{65559$\pm$3265(7.5)}} & \bf{\textit{65559$\pm$3265(7.5)}} \\ 
            & 4 & 36424$\pm$10(6) & \bf{\textit{29498$\pm$240(3.0)}} & \bf{\textit{30262$\pm$280(4.6)}} & \bf{\textit{30306$\pm$339(4.4)}} & \bf{\textit{28983$\pm$133(1.6)}} & \bf{\textit{28930$\pm$131(1.4)}} & \bf{\textit{59740$\pm$4565(7.5)}} & \bf{\textit{59740$\pm$4565(7.5)}} \\ 
            & 5 & \bf{\textit{32715$\pm$2377(4.0)}} & \bf{\textit{31524$\pm$348(3.7)}} & \bf{\textit{32271$\pm$345(5.2)}} & \bf{\textit{32185$\pm$417(5.1)}} & \bf{\textit{30444$\pm$167(1.4)}} & \bf{\textit{30437$\pm$160(1.6)}} & \bf{\textit{61165$\pm$3506(7.5)}} & \bf{\textit{61165$\pm$3506(7.5)}} \\ 
            \multirow{5}{*}{S3-C} & 1 & \bf{\textit{45906$\pm$0(5.0)}} & 47919$\pm$1147(6.0) & \bf{\textit{40932$\pm$333(3.4)}} & \bf{\textit{40930$\pm$425(3.6)}} & \bf{\textit{39026$\pm$146(1.4)}} & \bf{\textit{39049$\pm$154(1.6)}} & \bf{\textit{69270$\pm$3741(7.5)}} & \bf{\textit{69270$\pm$3741(7.5)}} \\ 
            & 2 & \bf{\textit{58082$\pm$263(5.0)}} & 58770$\pm$195(6.0) & \bf{\textit{51538$\pm$346(3.6)}} & \bf{\textit{51445$\pm$366(3.4)}} & \bf{\textit{49131$\pm$169(1.5)}} & \bf{\textit{49157$\pm$207(1.5)}} & \bf{\textit{78938$\pm$4233(7.5)}} & \bf{\textit{78938$\pm$4233(7.5)}} \\ 
            & 3 & \bf{\textit{45802$\pm$74(5.2)}} & 45868$\pm$21(5.8) & \bf{\textit{38727$\pm$341(3.4)}} & 38925$\pm$241(3.6) & \bf{\textit{36791$\pm$118(1.6)}} & \bf{\textit{36730$\pm$94(1.4)}} & \bf{\textit{67602$\pm$3362(7.5)}} & \bf{\textit{67602$\pm$3362(7.5)}} \\ 
            & 4 & 41011$\pm$1(6) & \bf{\textit{40879$\pm$115(5)}} & \bf{\textit{35457$\pm$319(3.5)}} & \bf{\textit{35339$\pm$338(3.5)}} & \bf{\textit{33663$\pm$89(1.7)}} & \bf{\textit{33653$\pm$91(1.3)}} & \bf{\textit{61607$\pm$2949(7.5)}} & \bf{\textit{61607$\pm$2949(7.5)}} \\ 
            & 5 & \bf{\textit{49309$\pm$114(5.5)}} & \bf{\textit{46793$\pm$3817(4.8)}} & \bf{\textit{42843$\pm$488(3.9)}} & \bf{\textit{42764$\pm$568(3.7)}} & \bf{\textit{40153$\pm$107(1.6)}} & \bf{\textit{40143$\pm$132(1.4)}} & \bf{\textit{72621$\pm$3858(7.5)}} & \bf{\textit{72621$\pm$3858(7.5)}} \\ 
            \multirow{5}{*}{S4-A} & 1 & 24849$\pm$1470(6) & \bf{\textit{21105$\pm$170(2.9)}} & \bf{\textit{22336$\pm$255(4.5)}} & \bf{\textit{22385$\pm$332(4.5)}} & \bf{\textit{20730$\pm$131(1.6)}} & \bf{\textit{20699$\pm$85(1.5)}} & \bf{\textit{53513$\pm$3603(7.5)}} & \bf{\textit{53513$\pm$3603(7.5)}} \\ 
            & 2 & \bf{\textit{24693$\pm$1073(3.1)}} & 24947$\pm$322(4.0) & \bf{\textit{26177$\pm$477(5.4)}} & \bf{\textit{26175$\pm$352(5.5)}} & \bf{\textit{23967$\pm$134(1.6)}} & \bf{\textit{23920$\pm$96(1.4)}} & \bf{\textit{59023$\pm$3019(7.5)}} & \bf{\textit{59020$\pm$3023(7.5)}} \\ 
            & 3 & 28262$\pm$12(6) & \bf{\textit{22203$\pm$286(3)}} & \bf{\textit{23181$\pm$356(4.4)}} & \bf{\textit{23091$\pm$319(4.5)}} & \bf{\textit{21661$\pm$67(1.6)}} & \bf{\textit{21638$\pm$68(1.4)}} & \bf{\textit{49301$\pm$3606(7.5)}} & \bf{\textit{49325$\pm$3575(7.5)}} \\ 
            & 4 & 28294$\pm$261(4.6) & \bf{\textit{27786$\pm$653(3.1)}} & \bf{\textit{28551$\pm$420(5.0)}} & \bf{\textit{28636$\pm$379(5.4)}} & \bf{\textit{26885$\pm$82(1.7)}} & \bf{\textit{26856$\pm$103(1.3)}} & \bf{\textit{62334$\pm$3972(7.5)}} & \bf{\textit{62334$\pm$3972(7.5)}} \\ 
            & 5 & \bf{\textit{28727$\pm$249(3.0)}} & 28989$\pm$389(3.6) & \bf{\textit{30571$\pm$479(5.5)}} & \bf{\textit{30693$\pm$441(5.5)}} & \bf{\textit{28421$\pm$139(1.6)}} & \bf{\textit{28440$\pm$125(1.8)}} & \bf{\textit{64233$\pm$5147(7.5)}} & \bf{\textit{64242$\pm$5138(7.5)}} \\ 
            \multirow{5}{*}{S4-B} & 1 & 47887$\pm$173(6) & \bf{\textit{40827$\pm$2573(3.3)}} & \bf{\textit{41301$\pm$476(4.4)}} & \bf{\textit{41242$\pm$464(4.3)}} & \bf{\textit{38564$\pm$117(1.6)}} & \bf{\textit{38549$\pm$139(1.4)}} & \bf{\textit{70139$\pm$4037(7.5)}} & \bf{\textit{70144$\pm$4029(7.5)}} \\ 
            & 2 & 43677$\pm$0(5.6) & \bf{\textit{43024$\pm$2961(4.7)}} & \bf{\textit{40667$\pm$448(3.7)}} & \bf{\textit{40804$\pm$492(4.0)}} & \bf{\textit{38055$\pm$128(1.5)}} & \bf{\textit{38061$\pm$172(1.5)}} & \bf{\textit{70219$\pm$3947(7.5)}} & \bf{\textit{70219$\pm$3947(7.5)}} \\ 
            & 3 & \bf{\textit{50458$\pm$193(3.4)}} & \bf{\textit{50440$\pm$239(3.2)}} & \bf{\textit{52701$\pm$547(5.6)}} & \bf{\textit{52632$\pm$519(5.4)}} & \bf{\textit{50091$\pm$210(1.6)}} & \bf{\textit{50103$\pm$151(1.8)}} & \bf{\textit{83768$\pm$3444(7.5)}} & \bf{\textit{83768$\pm$3444(7.5)}} \\ 
            & 4 & 34946$\pm$288(5.6) & \bf{\textit{33974$\pm$1347(3.2)}} & \bf{\textit{34526$\pm$317(4.4)}} & \bf{\textit{34478$\pm$318(4.7)}} & \bf{\textit{32820$\pm$95(1.5)}} & \bf{\textit{32864$\pm$148(1.5)}} & \bf{\textit{62765$\pm$3949(7.5)}} & \bf{\textit{62765$\pm$3949(7.5)}} \\ 
            & 5 & \bf{\textit{38698$\pm$4308(4.5)}} & \bf{\textit{38901$\pm$3107(4.8)}} & \bf{\textit{36166$\pm$349(4.4)}} & \bf{\textit{36195$\pm$289(4.4)}} & \bf{\textit{34049$\pm$163(1.6)}} & \bf{\textit{34016$\pm$143(1.4)}} & \bf{\textit{65365$\pm$4152(7.5)}} & \bf{\textit{65369$\pm$4147(7.5)}} \\ 
            \multirow{5}{*}{S4-C} & 1 & 42365$\pm$10(5.7) & \bf{\textit{42128$\pm$314(5.3)}} & \bf{\textit{35903$\pm$393(3.5)}} & \bf{\textit{35922$\pm$456(3.5)}} & \bf{\textit{33866$\pm$168(1.5)}} & \bf{\textit{33892$\pm$143(1.5)}} & \bf{\textit{61633$\pm$3417(7.5)}} & \bf{\textit{61633$\pm$3417(7.5)}} \\ 
            & 2 & \bf{\textit{49614$\pm$26(5)}} & 49797$\pm$74(6) & \bf{\textit{44052$\pm$394(3.4)}} & \bf{\textit{44087$\pm$340(3.6)}} & \bf{\textit{41842$\pm$170(1.4)}} & \bf{\textit{41878$\pm$204(1.6)}} & \bf{\textit{71240$\pm$3779(7.5)}} & \bf{\textit{71240$\pm$3779(7.5)}} \\ 
            & 3 & 50103$\pm$0(6) & \bf{\textit{49263$\pm$282(5)}} & \bf{\textit{41426$\pm$369(3.6)}} & \bf{\textit{41299$\pm$442(3.4)}} & \bf{\textit{39367$\pm$173(1.4)}} & \bf{\textit{39386$\pm$133(1.6)}} & \bf{\textit{71625$\pm$2776(7.5)}} & \bf{\textit{71625$\pm$2776(7.5)}} \\ 
            & 4 & 51311$\pm$0(6) & \bf{\textit{50781$\pm$48(5)}} & \bf{\textit{44940$\pm$369(3.4)}} & \bf{\textit{45099$\pm$410(3.6)}} & \bf{\textit{42895$\pm$168(1.4)}} & \bf{\textit{42939$\pm$179(1.6)}} & \bf{\textit{72347$\pm$3396(7.5)}} & \bf{\textit{72347$\pm$3396(7.5)}} \\ 
            & 5 & 46067$\pm$54(6) & \bf{\textit{45495$\pm$105(5)}} & \bf{\textit{40747$\pm$390(3.6)}} & \bf{\textit{40751$\pm$340(3.4)}} & \bf{\textit{38984$\pm$120(1.4)}} & \bf{\textit{38994$\pm$79(1.6)}} & \bf{\textit{66381$\pm$3802(7.5)}} & \bf{\textit{66381$\pm$3802(7.5)}} \\

            \midrule
            \multicolumn{4}{l}{Statistical results over 120 instances below}  &  &  &  &  &  & \\
            \midrule
            \multicolumn{2}{c}{\begin{tabular}[c]{@{}c@{}}Avr.Ranking \end{tabular}} & 5.25                              & 4.78                              & 3.95                             & 4.00                             & 1.51                             & 1.51                             & 7.50    &  7.5                 \\ 
            \multicolumn{2}{c}{$\#$of 'win-draw-lose'}              & \multicolumn{2}{c}{43-18-59}                                          & \multicolumn{2}{c}{4-116-0}                                         & \multicolumn{2}{c}{1-118-1}                                         &    \multicolumn{2}{c}{0-120-0}                            \\
            \multicolumn{2}{c}{p-value}              & \multicolumn{2}{c}{0.01}                                          & \multicolumn{2}{c}{0.15}                                         & \multicolumn{2}{c}{0.93}                                         &    \multicolumn{2}{c}{0.95}                             \\
            \bottomrule
            \end{tabular}
            \end{adjustbox}
    \end{adjustbox}
    \begin{tablenotes} 
		\item The data of E1-A $\sim$ S2-C are omitted in the table due to the page limitation. The complete data has been attached \sm{as supplementary} material. 
     \end{tablenotes}
\end{table*}

\subsection{Analysis of GOFVT}\label{experiment3}
The proposed optimisation framework GOFVT is capable of generalising \rev{almost all} algorithms for static CARPs to optimise DCARP. To demonstrate its \rev{effectiveness and efficiency}, we have selected three classical meta-heuristic algorithms for static CARPs, namely RTS \cite{mei2009global}, \sm{ILMA \cite{mei2009improved} and MAENS \cite{tang2009memetic}}, and \sm{embedded them into the} GOFVT in our experiments to evaluate whether it is advantageous to make use of existing static CARP algorithms within the GOFVT framework.

A brief description of each algorithm is presented below.

\begin{itemize}
    \item \textit{\textbf{RTS}} \cite{mei2009global}: A global repair operator which is \sm{embedded} in a tabu search algorithm (TSA \cite{brandao2008deterministic}). Source code is available \rev{online}\footnote{https://meiyi1986.github.io/publication/mei-2009-global/code.zip}.
    \item \textit{\textbf{ILMA}} \cite{mei2009improved}: An improved version of Lacomme’s memetic algorithm (LMA) \cite{lacomme2005evolutionary}. For our experiments, we implemented ILMA\footnote{\rev{https://github.com/HawkTom/Dynamic-CARP}} ourselves \sm{according to the details given in \cite{mei2009improved}}.
    \item \textit{\textbf{MAENS}} \cite{tang2009memetic}: A memetic algorithm with a merge-split operator. Source code is available \rev{online}\footnote{https://meiyi1986.github.io/publication/tang-2009-memetic/code.zip}.
\end{itemize}

The newly generated algorithms for optimising DCARP are denoted \rev{as} VT-RTS, VT-ILMA and VT-MAENS. We use these acronyms to denote GOFVT with the restart strategy. When using the sequence transfer strategy, they are denoted \rev{as} VTtr-RTS, VTtr-ILMA, VTtr-MAENS. 

\rev{To demonstrate the efficiency and robustness of our proposed framework, we use parameters as given in their original papers} \cite{mei2009global,tang2009memetic,mei2009improved}. 
In order to \rev{compare} the restart and sequence transfer strategies, a DCARP scenario consisting of 5 DCARP instances \sm{has been} generated for each \rev{static map in \textit{egl} benchmark set} in our experiments. A new DCARP instance is generated by executing the obtained best solution of the previous DCARP instance in our simulation system.

We have also embedded MASDC into GOFVT and generated two algorithms as VT-MASDC and VTtr-MASDC\rev{, respectively}. The results of above eight algorithms are presented in Table \ref{e3}. \sm{The values in each cell} represent the `mean$\pm$std' and the average ranking (in the brackets) \rev{among 8 algorithms} in terms of the average total cost over 25 independent runs.
The bold values represent the better result between \rev{the} restart strategy and \rev{the} sequence transfer strategy for each algorithm on a DCARP instance under the Wilcoxon signed-rank test with a significance level of 0.05. Rows where both \rev{the restart strategy and sequence transfer strategy} are in bold \rev{imply that} these two strategies are not significantly different under the hypothesis test. The summaries of win-draw-lose of \sm{the} restart strategy versus the sequence transfer strategy on all DCARP instances are listed in the penultimate row. We have also conducted the Wilcoxon signed-rank test with a significance level of 0.05 for the average total cost of \sm{the} restart strategy and sequence transfer strategy of each meta-heuristic algorithm on all instances, and the \rev{p-values} associated with each meta-heuristic algorithm are listed in the last row of Table \ref{e3}.

We \sm{conclude} that the restart strategy is significantly different from the sequence transfer strategy when embedded in RTS (0.01 $<$ 0.05), but \sm{both} strategies obtain \sm{a} similar performance when embedded in ILMA (0.15 $>$ 0.05), MAENS (0.93 $>$ 0.05) and MASDC (0.95 $>$ 0.05). This is mainly because RTS is an individual-based meta-heuristic algorithm while ILMA, MAENS and MASDC are population-based algorithms. An individual-based algorithm only uses one solution \rev{during} optimisation. The solution generated by the sequence transfer strategy will be the only initial solution in the individual-based algorithm and \sm{therefore} significantly influences the optimisation result. In contrast, the population-based algorithm contains a population during the optimisation so that it depends much less on a single \rev{transferred} solution.

From the statistical test result and the \sm{number} of `win-draw-lose' of all DCARP instances for the individual-based algorithm, i.e. RTS, we \rev{can conclude that the performance of the sequence transfer strategy is better than the restart strategy}. The \sm{efficiency} depends on how much information is \rev{transferred} from the previous best solution. For a CARP solution, the most critical information is the sequence of tasks of each route. Therefore, if each route has many tasks left, and these remaining tasks can also maintain the tasks' sequence of the best solution in the previous instance, the \rev{transferred} solution will be \rev{of} high quality. The sequence transfer strategy fixes the remaining tasks' position to maintain the sequence information. Then, new tasks are inserted into the sequence to construct a new initial solution. If an outside vehicle has only a few remaining tasks, most tasks in the new schedule will be the new tasks so that the new schedule is unlikely to benefit \rev{much} from the previous best schedule.

In contrast, if there are many remaining tasks for an outside vehicle, the order of remaining tasks will be maintained in the new sequence of tasks, and the Ulusoy's split will still assign them to an outside vehicle. \sm{Consider} the following two remaining task sequences as an example:
\begin{equation*}
    \begin{aligned}
       S_1: & \rev{(} v_0, vt_1, t_1, v_0, vt_2, t_2, vt_3, t_3, v_0 \rev{)}, \\
       S_2: & \rev{(} v_0, vt_1, t_1, t_2, t_3,t_4, t_5, v_0, vt_2, t_6, t_7, t_8,t_9, t_{10}, t_{11}, v_0 \rev{)}. 
    \end{aligned}
\end{equation*}
$S_1$ has three outside vehicles with one remaining task for each vehicle, and  $S_2$ has two outside vehicles with 5 and 6 remaining tasks, respectively. The task sequences in both remaining task sequences are the same as those of the previous instance's best schedule. After greedy insertion of a set of new tasks, the final task sequence served by outside vehicles in $S_2$ will be more similar to the schedule in the previous instance's best solution. As a result, the second scenario is more likely to obtain a high-quality \rev{transferred} solution.

\begin{figure}[!htpb]
    \centering
    \includegraphics[width=0.8\linewidth]{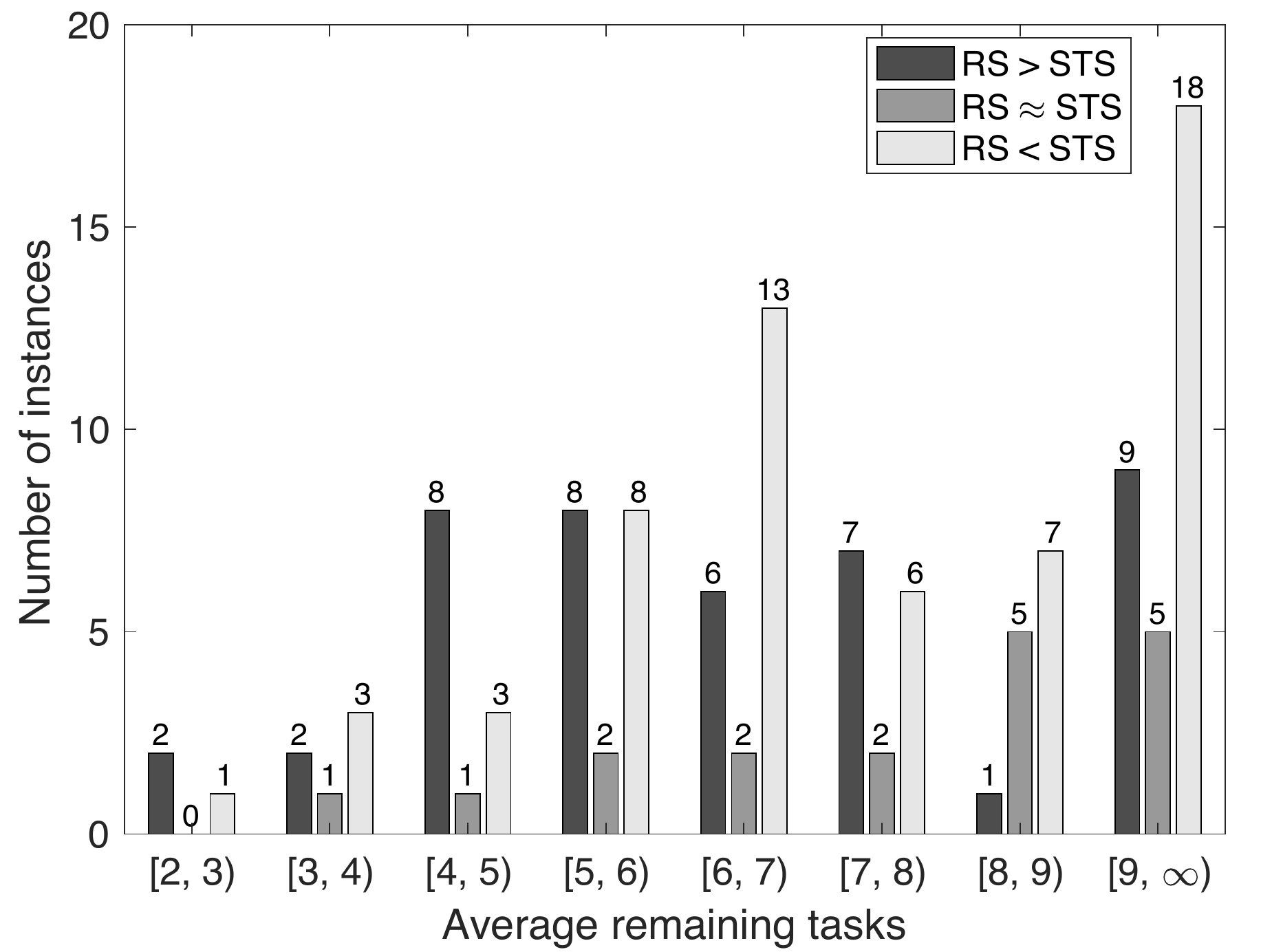}
    \caption{The number of instances for the restart strategy (RS) wins over ($\succ$), draws against ($\approx$) and loses to ($\prec$) the sequence transfer strategy (STS) for different ranges of average remaining tasks.} \label{rov1}
\end{figure}

We have calculated the average remaining tasks for outside vehicles on all DCARP instances in our experiment \rev{in Figure \ref{rov1}}. \rev{We can conclude that as} the average remaining tasks for each outside vehicle increases, the sequence transfer strategy benefits more from the optimisation experience in the previous instance and outperforms the restart strategy with a higher probability.

\begin{figure}[!htpb]
    \centering
    \includegraphics[width=\linewidth]{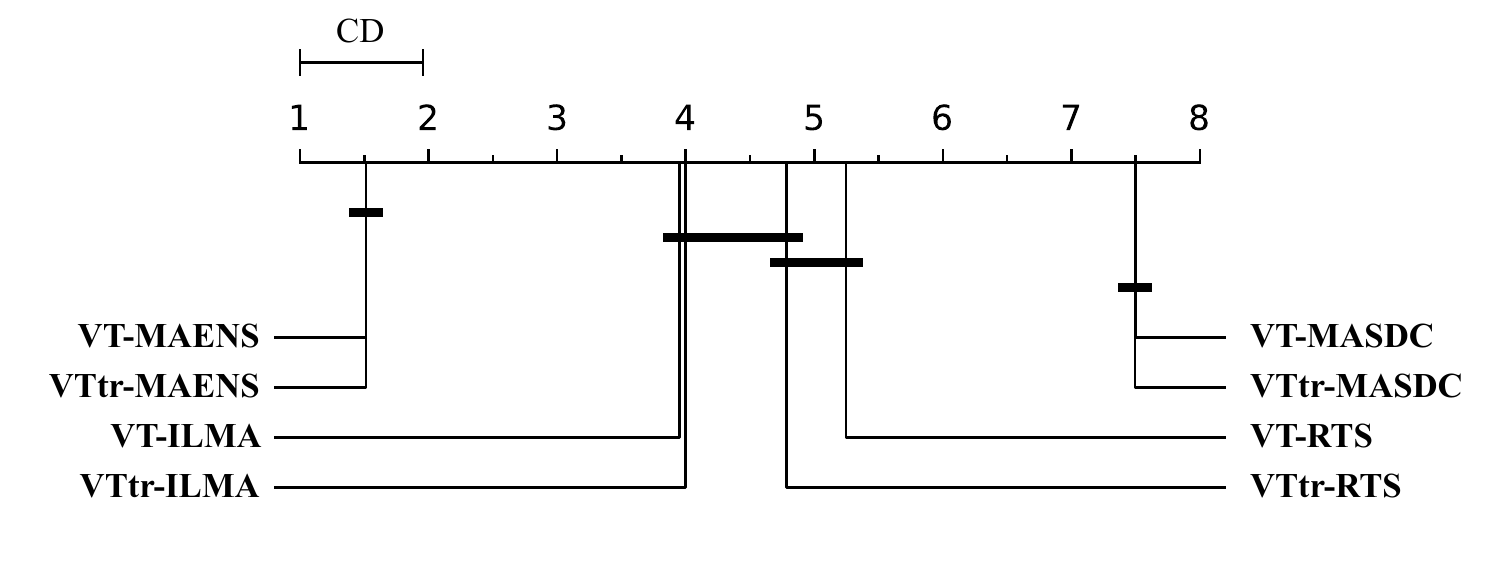}
    \caption{Critical difference diagram for the comparison of 8 algorithms against each other on \textit{egl} with Friedman test and Nemenyi test. Groups of algorithms \sm{which are} not significantly different at the level of 0.05 are connected.} \label{cd}
\end{figure}

We also compared \sm{the} rankings of each algorithm over 25 independent runs on each DCARP instance, which is presented in each cell's bracket in Table \ref{e3}. The overall average rankings of each algorithm instance over 120 DCARP instances are summarised in the row of `Avr. Ranking' in Table \ref{e3}. The Friedman test with a significance level of 0.05 was carried out to compare the ranking of eight algorithms across problem instances, leading to a p-value of 8.69e-159. This indicates that at least one pair of algorithms are not equivalent to each other. We then perform Nemenyi posthoc tests to identify which algorithms perform significantly different. The critical difference diagram is presented in Figure \ref{cd} where the value of critical difference is 0.96 \cite{demvsar2006statistical}. \rev{We can conclude from these results} that the restart and sequence transfer strategies obtained almost the same performance for population-based algorithms (MAENS and ILMA) while the sequence transfer strategy slightly outperformed the restart strategy in the individual-based algorithm (RTS) in our experiments.

Furthermore, \rev{MAENS} obtains the best \rev{overall} result, and RTS is the worst among the three meta-heuristic algorithms which are originally designed for static CARP. However, all of them significantly outperform the \rev{state-of-the-art} dynamic algorithm, i.e. VT-MASDC and VTtr-MASDC. Recall that VT-MASDC was shown to outperform MASDC in Section \ref{experiment2}. Therefore, we \rev{can} conclude that not only the proposed framework generalises the existing algorithms for static CARPs to solve DCARPs \sm{but also that} the constructed algorithm maintains its superior performance when optimising a DCARP instance.

\rev{To evaluate the impact of parameter configuration of algorithms on our final conclusion, we have used SMAC \cite{hutter2011sequential} to obtain the best parameters for MAENS and MASDC, and then compared meta-heuristic algorithms with tuned parameters on additional 72 DCARP instances. The results are presented in Table VII of the supplementary material. They show clearly that MASDC with tuned parameters still performed worse than MAENS with the default or tuned parameters. MAENS with tuned parameters preformed similar to MAENS with default parameters in our experiments. In short, different parameter settings for the meta-heuristic algorithms do not change our conclusions.}

\section{Conclusion and future work}\label{conclusion}
In this paper, we \rev{studied} the dynamic capacitated arc routing problem (DCARP), in which dynamic events, such as road closure, added tasks, etc., occur during the vehicles' service. First, a mathematical formulation of DCARP was provided for the first time in the literature. Then, we designed a simulation system as the research platform for DCARP. Unlike the existing benchmark generator, our simulation system generates DCARP instances from a given map \sm{in a way} that \rev{is more realistic and closer to the real world. Our simulator also facilitates the use of existing algorithms for static CARP to be used for DCARP.} The events simulated in our system enable existing benchmark maps from static CARP to be adopted in \rev{dynamic} scenarios much closer to the real world, where dynamic events happen while a solution is being deployed, i.e., while vehicles are serving \rev{in} the map. Given the mathematical model and the simulation system, we \rev{have} proposed a generalised optimisation framework which can generalise algorithms for static CARPs to optimise DCARPs. In our framework, we proposed a virtual-task strategy that constructs a virtual task between the stop location and the depot, to make all outside vehicles virtually return to the depot, which \sm{tremendously} simplified the optimisation for DCARP. As a result, the DCARP instance was converted into a virtual `static' instance so that algorithms for static CARP can solve it. 

Two \rev{initialisation} strategies were \rev{adopted} in \rev{our} generalised optimisation framework: the restart and sequence transfer strategies. The restart strategy completely restarts the optimisation algorithm \rev{at random} when there is a new DCARP instance. The sequence transfer strategy maintains the sequence of remaining tasks to the new DCARP instance and greedily inserts the new tasks into the remaining sequence, to transfer the information of task sequence from the previous optimisation experience.

In our computational study, the necessity of directly optimising DCARP together with outside vehicles was first demonstrated by comparing the virtual-task strategy with a return-first strategy. The results \sm{indicated} that it would be more efficient to optimise the DCARP instance by using the virtual-task strategy when \sm{the} outside vehicles' remaining capacities were \sm{sufficiently large} to serve more tasks. Then, the efficiency of the virtual-task strategy was demonstrated by embedding it into an existing algorithm and comparing it to the original version of the existing algorithm. Finally, \rev{our} generalised optimisation framework's efficiency was analysed by integrating a set of optimisation algorithms that were designed for static CARPs, and the constructed algorithms performed significantly better than \rev{state-of-the-art} algorithms for DCARP.

\rev{In this paper, we have demonstrated that it is effective to solve DCARP instances by transforming them into static instances using the VT strategy and using our proposed GOFVT framework. The influence of the type and degree of dynamic events on individual optimisation algorithms was not investigated. We will further investigate the influence of different dynamic events on optimisation algorithms and design a benchmark including DCARP instances with different dynamic characteristics.}
\rev{Furthermore}, the current sequence transfer strategy only uses the optimisation experience belonging to the instance of the previous optimisation. \rev{It would be} valuable to utilise all previous search experience taken from the whole \rev{DCARP} scenario. Finally, it is desirable to test \rev{our} proposed framework \rev{further} with large scale \rev{CARP} instances and real world applications in the future.

\section*{Acknowledgements}
Hao Tong gratefully acknowledges the financial support from Honda Research Institute Europe (HRI-EU). Part of this work was done while the first author was a visiting PhD student at SUSTech. This work was supported by the Honda Research Institute Europe (HRI-EU), the Guangdong Provincial Key Laboratory (Grant No. 2020B121201001), the Program for Guangdong Introducing Innovative and Entrepreneurial Teams (Grant No. 2017ZT07X386), Shenzhen Science and Technology Program (Grant No. KQTD2016112514355531), the Guangdong Basic and Applied Basic Research Foundation (Grant No. 2021A1515011830) and the Research Institute of Trustworthy Autonomous Systems (RITAS). 

\bibliographystyle{IEEEtran}
\bibliography{reference}

\clearpage
\newpage


\section*{Supplementary material}

**IV. Experimental results and analysis for investigating the influence of different dynamic scenarios on optimisation algorithm.**

TABLE VI

RESULTS OF MANES OVER 25 INDEPENDENT RUNS ON DIFFERENT DYNAMIC SCENARIOS WITH DIFFERENT EVENT DEGREE ON MAP $egl - S1 - A$. THE EVENT DEGREE DENOTES THE PERCENTAGE OF ARCS THAT THEIR COST OR DEMAND CHANGES.

| Scenario Description | Instances' Index | Event Degree($\mu$) | Cost | performance |
|---|---|---|---|---|
| | 1 | 10.00% | 7745±33 | 0.444 |
| | 2 | 20.00% | 6566±21 | 0.557 |
| | 3 | 30.00% | 8626±39 | 0.506 |
| | 4 | 40.00% | 8370±9 | 0.523 |
| | 5 | 50.00% | 8095±21 | 0.474 |
| Congestion: Added Roads' cost | 6 | 60.00% | 8339±20 | 0.503 |
| | 7 | 70.00% | 9849±43 | 0.512 |
| | 8 | 80.00% | 9418±36 | 0.529 |
| | 9 | 90.00% | 12618±44 | 0.475 |
| | 10 | 100.00% | 11332±25 | 0.497 |
| | 1 | 10.00% | 8653±54 | 0.508 |
| | 2 | 20.00% | 9124±75 | 0.455 |
| | 3 | 30.00% | 10043±48 | 0.445 |
| | 4 | 40.00% | 12312±71 | 0.454 |
| | 5 | 50.00% | 11101±56 | 0.430 |
| New Tasks | 6 | 60.00% | 11731±33 | 0.435 |
| | 7 | 70.00% | 9674±51 | 0.451 |
| | 8 | 80.00% | 9064±35 | 0.457 |
| | 9 | 90.00% | 9695±45 | 0.426 |
| | 10 | 100.00% | 10802±50 | 0.436 |
| | 1 | 10.00% | 7710±45 | 0.473 |
| | 2 | 20.00% | 10096±37 | 0.499 |
| | 3 | 30.00% | 13393±53 | 0.420 |
| | 4 | 40.00% | 13802±37 | 0.427 |
| Congestion and New Tasks | 5 | 50.00% | 15848±78 | 0.443 |
| | 6 | 60.00% | 13328±51 | 0.453 |
| | 7 | 70.00% | 14719±58 | 0.463 |
| | 8 | 80.00% | 17568±92 | 0.456 |
| | 9 | 90.00% | 20278±90 | 0.433 |
| | 10 | 100.00% | 16732±179 | 0.441 |

An analysis of the impact of the generators on the performance of the algorithms would enable to identify which parameter values have a more severe impact on the performance of the underlying meta-heuristic being adopted with our framework.

We have many parameters in our simulator, but all of them mainly affect the cost and demand of an arc. Therefore, we create two key scenarios, i.e. congestion and added tasks, that related to most of these parameters. The congestion related to parameters $p_{event}, p_{road}, p_{bdrr}, p_{crr}, p_{crbb}$ and the added tasks related to $p_{icd}, p_{add}$. Then, different parameter setting will influence the degree of dynamic events. For example, smaller $p_{event}$ causes fewer congestion and smaller $p_{add}$ results fewer new tasks. Therefore, we generated instances with different degree $\mu$ ($\mu = 0.1, 0.2, ..., 1.0$) of congestion of congestion and new tasks.

We have chosen to investigate the impact of these parameters using MAENS. The final results are presented on Table VI. In our experiments, we select $\mu$ percent of all arcs to increase its cost for congestion scenario, and select $\mu$ percent of all available zero-demand arcs as the new tasks for added tasks scenario. From our result, we plot curves in Figure 1 in this response letter, and the performance is calculated by the following equation. As shown in Figure 1, the new added tasks influence MAENS's performance much more than congestion, because it increase the problem's dimension.

We calculated the performance using the formula below:

$$performance = \frac{Cost_{initial\_best\_solution} - Cost_{final\_best\_solution}}{Cost_{initial\_best\_solution} - Cost_{lower\_bound}}$$

Figure 1 shows the performance obtained for different event degrees and types of change. Table 3 in the appendix also shows these results in table format. We can see from Figure 1 that new added tasks influence MAENS' performance much more than congestion. This is reasonable, as it increases the problem's dimension.

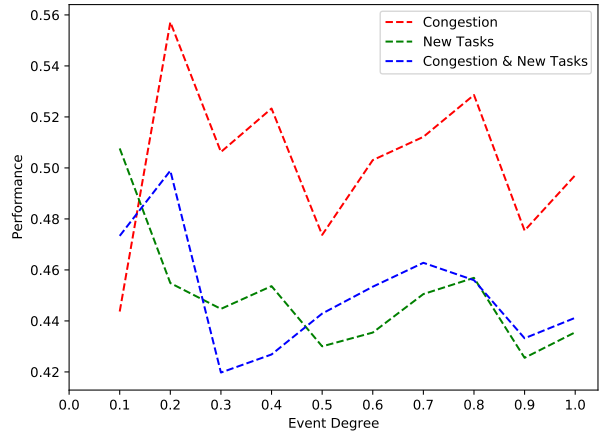

Fig. 1. The influence of different degree of dynamic events, i.e. congestion, new tasks, in DCARP on MAENS algorithms.

## V. Experimental results for investigate the influence of parameter configuration on algorithm's performance.

We have investigated the following parameters for MAENS and MASDC, as listed in Table VII. We tuned these two algorithms' parameter by SMAC. The default and tuned parameters for these two algorithms are also presented in Table VII.

TABLE VII
The default and tuned parameter setting (By SMAC) for algorithms MAENS and MASDC.

| Algorithm | Default | Tuned |
|---|---|---|
| MANES | $opsize = 6 * psize, p = 2, P_{ls} = 0.2$ | $opsize = 6 * psize, p = 2, P_{ls} = 0.334$ |
| MASDC | $P_{ls} = 0.2$ | $P_{ls} = 0.250$ |

We compare algorithm MAENS with default paramter setting with MAENS with tuned parameter setting (MAENS*), and with MASDC with tuned paremeter setting*, whose results are presented in TableVIII. From the table, even though the MASDC with tuned parameter setting, its performance still much worse than the MAENS with default setting. On the other hand, the MAENS performs almost same with MAENS*, which indicates the parameter setting has no significant influence on MAENS. As a sequence, the tuned parameter settings almost have no impact to our paper's conclusion, that the proposed framework with existing algorithms for static CARP obtains a better performance than the state-of-the-art DCARP algorithms.

TABLE VIII
Results of MAENS, MAENS with tuned parameter setting (MAENS*) and MASDC with tuned parameter setting (MASDC*) on 72 DCARP instances from the *egl* dataset. The value in each cell represents "Mean ± Std" over 25 independent runs and the bold ones denote the better result on the DCARP instance based on the Wilcoxon signed-rank test with the level of significance 0.05. The last row summaries the number of win-draw-lose of two compared algorithms.

| Map | Instance | MAENS | MAENS* | MAENS | MASDC* |
|---|---|---|---|---|---|
| E1-A | 1 | 9914±95 | 9805±27 | 9914±95 | 21986±5320 |
| | 2 | 7893±58 | 7818±48 | 7893±58 | 22154±5364 |
| | 3 | 9728±47 | 9693±41 | 9728±47 | 28775±752 |
| E1-B | 1 | 8616±44 | 8586±41 | 8616±44 | 19451±4687 |
| | 2 | 10891±82 | 10877±89 | 10891±82 | 24996±6022 |
| | 3 | 10920±47 | 10899±33 | 10920±47 | 29250±844 |
| E1-C | 1 | 10814±20 | 10812±21 | 10814±20 | 23182±5582 |
| | 2 | 13102±53 | 13050±50 | 13102±53 | 25227±6066 |
| | 3 | 12895±28 | 12882±29 | 12895±28 | 27451±753 |
| E2-A | 1 | 8731±63 | 8711±58 | 8731±63 | 20814±5010 |
| | 2 | 10472±58 | 10450±52 | 10472±58 | 23230±5619 |
| | 3 | 9811±50 | 9808±58 | 9811±50 | 29579±904 |
| E2-B | 1 | 10641±19 | 10636±2 | 10641±9 | 23279±5623 |
| | 2 | 11556±45 | 11527±47 | 11556±45 | 24140±5806 |
| | 3 | 13406±23 | 13408±18 | 13406±23 | 28792±579 |
| E2-C | 1 | 17131±33 | 17110±24 | 17131±33 | 29573±7113 |
| | 2 | 15158±146 | 15032±126 | 15158±146 | 29809±7147 |
| | 3 | 15738±5 | 15737±1 | 15738±5 | 30455±547 |
| E3-A | 1 | 10840±64 | 10841±57 | 10840±64 | 24553±5901 |
| | 2 | 11931±98 | 11894±104 | 11931±98 | 26719±6463 |
| | 3 | 10416±56 | 10421±46 | 10416±56 | 26737±892 |
| E3-B | 1 | 13999±0 | 14007±21 | 13999±0 | 27731±6708 |
| | 2 | 17311±28 | 17291±19 | 17311±28 | 32671±7856 |
| | 3 | 20793±59 | 20810±45 | 20793±59 | 38550±789 |
| E3-C | 1 | 18423±84 | 18441±89 | 18423±84 | 30396±7291 |
| | 2 | 19885±8 | 19871±21 | 19885±8 | 35314±8468 |
| | 3 | 16665±75 | 16631±89 | 16665±75 | 34201±1192 |
| E4-A | 1 | 12162±45 | 12158±36 | 12162±45 | 26422±6352 |
| | 2 | 14583±62 | 14597±73 | 14583±62 | 29559±7117 |
| | 3 | 12492±53 | 12485±58 | 12492±53 | 30088±741 |
| E4-B | 1 | 15626±40 | 15606±63 | 15626±40 | 29242±6998 |
| | 2 | 16541±49 | 16512±53 | 16541±49 | 32618±7802 |
| | 3 | 15682±57 | 15645±30 | 15682±57 | 31365±709 |
| E4-C | 1 | 17139±37 | 17113±68 | 17139±37 | 30959±7467 |
| | 2 | 19914±53 | 19892±40 | 19914±53 | 34818±8382 |
| | 3 | 20338±61 | 20287±73 | 20338±61 | 37959±668 |

| Map | Instance | MAENS | MAENS* | MAENS | MASDC* |
|---|---|---|---|---|---|
| S1-A | 1 | 14238±55 | 14228±50 | 14238±55 | 39105±9429 |
| | 2 | 20171±86 | 20143±65 | 20171±86 | 48755±11652 |
| | 3 | 23748±152 | 23796±202 | 23748±152 | 66086±1676 |
| S1-B | 1 | 21979±78 | 22036±70 | 21979±78 | 46286±11062 |
| | 2 | 20446±104 | 20418±78 | 20446±104 | 49206±11889 |
| | 3 | 30294±104 | 30370±110 | 30294±104 | 78355±1888 |
| S1-C | 1 | 25597±74 | 25600±87 | 25597±74 | 43925±10507 |
| | 2 | 32961±95 | 33076±180 | 32961±95 | 63290±15176 |
| | 3 | 34557±100 | 34664±122 | 34557±100 | 73934±1160 |
| S2-A | 1 | 21629±68 | 21635±72 | 21629±68 | 52556±12588 |
| | 2 | 26376±113 | 26398±137 | 26376±113 | 64336±15473 |
| | 3 | 20063±132 | 20121±84 | 20063±132 | 59546±1352 |
| S2-B | 1 | 24646±131 | 24736±114 | 24646±131 | 51048±12323 |
| | 2 | 25977±166 | 25949±141 | 25977±166 | 61779±14827 |
| | 3 | 27215±106 | 27260±136 | 27215±106 | 66500±1721 |
| S2-C | 1 | 33881±117 | 33924±172 | 33881±117 | 66018±15799 |
| | 2 | 43764±178 | 43807±207 | 43764±178 | 82736±19808 |
| | 3 | 39370±222 | 39487±177 | 39370±222 | 80966±1186 |
| S3-A | 1 | 19915±87 | 19948±56 | 19915±87 | 53690±12998 |
| | 2 | 24939±102 | 24961±116 | 24939±102 | 69849±16783 |
| | 3 | 29522±125 | 29603±131 | 29522±125 | 89771±2681 |
| S3-B | 1 | 27262±89 | 27339±79 | 27262±89 | 54242±13026 |
| | 2 | 36059±125 | 36045±111 | 36059±125 | 71379±17165 |
| | 3 | 33613±152 | 33682±148 | 33613±152 | 80332±1657 |
| S3-C | 1 | 36479±159 | 36548±180 | 36479±159 | 66757±16024 |
| | 2 | 36833±162 | 36859±129 | 36833±162 | 64195±15346 |
| | 3 | 36963±148 | 37011±110 | 36963±148 | 75722±1808 |
| S4-A | 1 | 23802±83 | 23817±118 | 23802±83 | 56578±13520 |
| | 2 | 24611±104 | 24629±120 | 24611±104 | 61565±14802 |
| | 3 | 26700±161 | 26722±107 | 26700±161 | 69200±2007 |
| S4-B | 1 | 32657±127 | 32707±195 | 32657±127 | 71871±17298 |
| | 2 | 29262±91 | 29286±99 | 29262±91 | 64618±15485 |
| | 3 | 29182±155 | 29248±169 | 29182±155 | 69618±1605 |
| S4-C | 1 | 41551±122 | 41551±122 | 41551±122 | 70663±16958 |
| | 2 | 39078±155 | 39095±199 | 39078±155 | 75359±17998 |
| | 3 | 45284±166 | 45314±136 | 45284±166 | 94029±1954 |

| | | | | | |
|---|---|---|---|---|---|
| # of w-d-l | | 2-62-8 | | 72-0-0 | |
| p-value | | 0.29 | | 1.66e-13 | |



\end{document}